\documentclass[manuscript,screen,review=false]{acmart}

\AtBeginDocument{%
  \providecommand\BibTeX{{%
    \normalfont B\kern-0.5em{\scshape i\kern-0.25em b}\kern-0.8em\TeX}}}

\setcopyright{acmcopyright}
\copyrightyear{2023}
\acmYear{2023}
\acmDOI{XXXXXXX.XXXXXXX}

\acmConference[Conference acronym 'XX]{Make sure to enter the correct
  conference title from your rights confirmation emai}{June 03--05,
  2018}{Woodstock, NY}
\acmPrice{15.00}
\acmISBN{978-1-4503-XXXX-X/18/06}

\usepackage{longtable}
\usepackage{adjustbox}
\usepackage{graphicx}
\usepackage{array}

\begin{document}

\title{Dense Video Captioning: A Survey of Techniques, Datasets and Evaluation Protocols}

\author{Iqra Qasim}
\email{iqra.qasim@uit.no}
\orcid{1234-5678-9012}
\email{webmaster@marysville-ohio.com}
\affiliation{%
  \institution{UiT The Arctic University of Norway}
  \streetaddress{Hansine Hansens Veg}
  \city{Tromso}
  \country{Norway}
  \postcode{9037}
}

\author{Alexander Horch}
\affiliation{%
 \institution{UiT The Arctic University of Norway}
 \streetaddress{Hansine Hansens Veg}
 \city{Tromso}
 \country{Norway}}
 \postcode{9037}

\author{Dilip K. Prasad}
\affiliation{%
 \institution{UiT The Arctic University of Norway}
 \streetaddress{Hansine Hansens Veg}
 \city{Tromso}
 \country{Norway}}
 \postcode{9037}
 \email{postmottak@uit.no}

\renewcommand{\shortauthors}{Qasim and Horsch, et al.}

\begin{abstract}
  Untrimmed videos have interrelated events, dependencies, context, overlapping events, object-object interactions, domain specificity, and other semantics that are worth highlighting while describing a video in natural language. Owing to such a vast diversity, a single sentence can only correctly describe a portion of the video. Dense Video Captioning (DVC) aims at detecting and describing different events in a given video. The term DVC originated in the 2017 ActivityNet challenge, after which considerable effort has been made to address the challenge. Dense Video Captioning is divided into three sub-tasks: (1) Video Feature Extraction (VFE), (2) Temporal Event Localization (TEL), and (3) Dense Caption Generation (DCG). This review aims to discuss all the studies that claim to perform DVC along with its sub-tasks and summarize their results. We also discuss all the datasets that have been used for DVC. Lastly, we highlight some emerging challenges and future trends in the field. 
\end{abstract}



\received{20 February 2007}
\received[revised]{12 March 2009}
\received[accepted]{5 June 2009}

\maketitle

\section{Introduction}
Interest in video content analysis has increased alongside the rapid expansion of large-scale video datasets made possible by the proliferation of video-sharing platforms like YouTube, Netflix, and Dailymotion, to name a few. More recently, developing 2D and 3D convolutional neural networks (CNNs) has sparked interest in studying static and dynamic visual media's encoding, captioning, and query-answering capabilities. However, accomplishing these tasks on long, unedited video significantly challenges computer vision. Dense video captioning aims to make a computer understand what is happening in a video and establish a relation between the video content and its meaningful natural language description. The capability of describing events in videos aids a variety of systems, including blind navigation, video searching, surveillance, medical image analysis, and automatic video subtitling.

The urge to detect captions on images and videos started in 1970 when researchers began working with images and video snippets containing captions. The art of displaying text on images and video transcribing the audio is called \textit{closed captioning}. To serve the consumers who are hard of hearing and to take part in technology improvement motivated researchers to develop some automatic caption detection systems \cite{845381, 646033}. They used binary edge maps, second-order statistical classifiers, Optical Character Recognition (OCR), video content analysis, automatic indexing systems for video data, and other such approaches for text-localization/extraction and visual content analysis. Studies formed in the 1970s laid the foundation of many modern computer vision techniques and algorithms, such as edge extraction and scene understanding. The National Captioning Institute \footnote{\label{NCI} \href{https://www.ncicap.org/}{National Captioning Institute (NCI)}} devised a real-time captioning solution in 1982 that could be applied to a live telecast. Until the end of the 1990s, three key visual content retrieval approaches had appeared: image feature extraction, word detection based on form and texture similarities, and video interpreting \cite{aigrain1996content}. Another significant breakthrough of that era was a real-time face recognition system from images and video with the help of eigenfaces to aid image and video analysis \cite{139758, liu2000face}. 

First decade of 21st century can be classified as a classical era of visual content analysis. Many breakthrough video processing, research, and understanding techniques evolved during this era. Some of the techniques include video retrieval using semantics \cite{hauptmann2008video}, video shot-boundary detection \cite{ hanjalic2002shot}, keyframe extraction \cite{yu2004multilevel}, cluster-based object trajectory acquisition \cite{hu2007semantic}, histogram-based object matching for video retrieval \cite{sivic2003video}, video data mining \cite{shyu2008video}, video annotation and retrieval \cite{ballan2010video}, video tagging with metadata \cite{aradhye2009video2text}, and clustering captions and videos \cite {huang2013multi}. A significant paradigm shift is worth mentioning when language-based postprocessing techniques \cite{guadarrama2013youtube2text, krishnamoorthy2013generating, Das_2013_CVPR}. Such techniques are referred to as two-stage pipelines / template-based / rule-based methods. Rule-based methods exploit certain rules/templates for generating captions by matching the attributes identified in videos. The current paradigm of caption generation works with seq-2-seq modeling of events and caption generation.

\begin{figure}[H]
  \centering
  \includegraphics[width=420px]{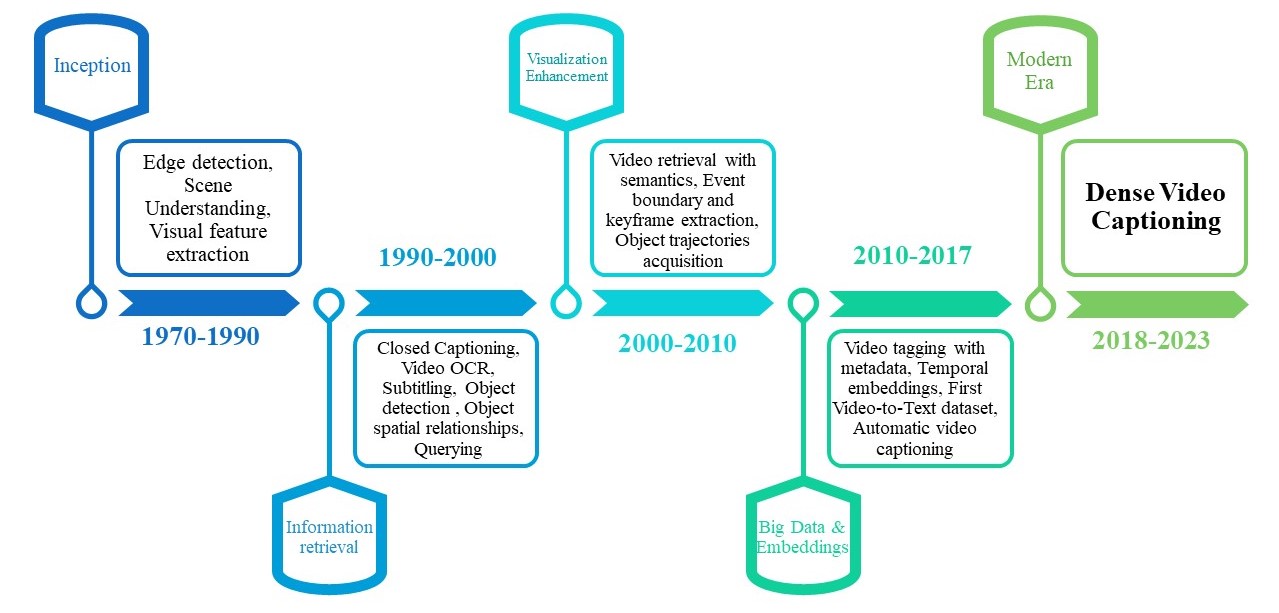}
  \caption{Advent and advances in visual data analysis techniques}
\end{figure}

Classical image captioning techniques such as \cite{ vinyals2015show} inspired modern video captioning techniques. These techniques first extract image features using CNN to generate a fixed-length vector representation of a video, which is then decoded into a sequence of words using RNNs. Since RNNs are unsuitable for long-range dependencies, they were replaced by LSTMs \cite{dai2015semi}. Nowadays, most dense video captioning techniques follow LSTMs and encoder-decoder architectures to generate event-dependent captions of long, untrimmed videos. The field of Video Content Analysis (VCA) became wider when researchers started video synopsis or video summarization of long videos \cite{9594911}.

Dense captioning is a step ahead of regular categorization of events in the video as it uses timestamp information. By dense, we mean capturing all details in a video that describe 'what's happening' inside a video. DVC was first addressed in \cite{Krishna_2017_ICCV}.

\begin{figure}
    \centering
    \includegraphics[width=350px]{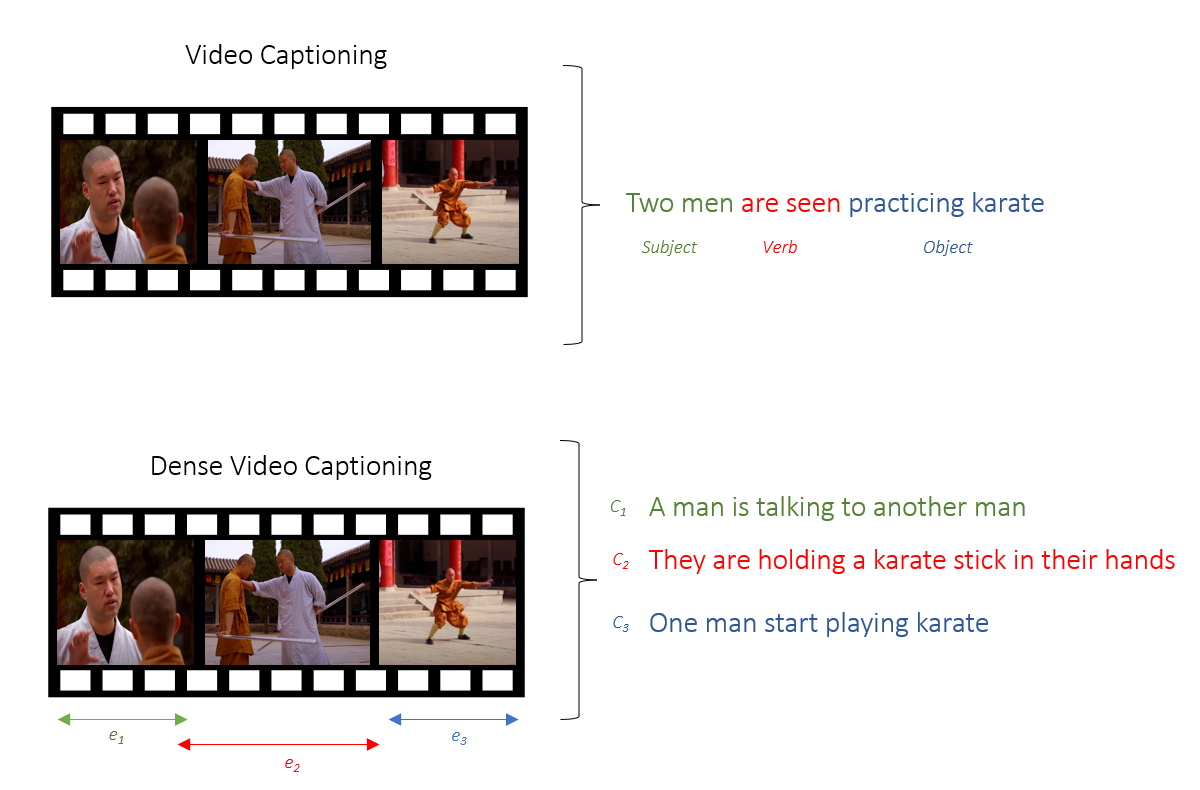}
    \caption{Video Captioning (VC) Vs Dense Video Captioning (DVC) task performance on a video. DVC generates more detailed captions for a given video while VC generates single sentence describing the video in general.}
    \label{fig:intro picture showing VC and DVC difference}
\end{figure}

\subsection{Motivation and contribution }
This survey aims to investigate the number of Dense Video Captioning methods that have evolved since the premise of the DVC and ActivityNet challenge. We aim to summarize all methods used for DVC video feature extraction, temporal event localization, and caption generation. A review still needs to participate in a systematic overview of the methods used for DVC. Hence, this survey is a timely contribution to the topic. The contribution of this survey is three-fold: (1) novel comprehensive techniques synthesis, (2) benchmark and comparative evaluation, and getting insights from literature and exploring future endeavors.

\subsection{Research Questions}
Our survey attempts to perform a thorough methodological overview of the methods involved in DVC as opposed to the existing surveys discussed in Table \ref{tab:survey-table}. The survey focuses on the following research questions: 
\begin{enumerate}
    \item What are the most effective and recently used methods for extracting features from the video datasets? 
    \item What techniques are being used for temporal event localization in videos for the purpose of DVC?
    \item What captioning strategies have been followed during the past 5 years (2018-2023)?
    \item What are the current trends in publishing ‘dense video captioning’? 
    \item What are expected future trends in dense event captioning with reference to open challenges? 
\end{enumerate}

\begin{table}[H]
\caption{A comprehensive directory of recent surveys in Video Captioning (2019-2023)}
\label{tab:survey-table}
\begin{tabular}{llll}
\hline
\textbf{Year} & \textbf{Title}                                                                                                                      & \textbf{Ref.}                  & \textbf{Publication Venue}                                                                                             \\ \hline
2019          & \begin{tabular}[c]{@{}l@{}}Video Description: A Survey of Methods,   \\ Datasets, and Evaluation Metrics\end{tabular}               & \cite{aafaq2019video}         & ACM computing surveys                                                                                                   \\ \hline
2020          & \begin{tabular}[c]{@{}l@{}}A Comprehensive Review on Recent Methods and  \\ Challenges of Video Description\end{tabular}            & \cite{singh2020comprehensive} & ACM                                                                                                                     \\ \hline
2021          & \begin{tabular}[c]{@{}l@{}}Exploring Video Captioning Techniques: \\ A Comprehensive Survey on Deep Learning Methods\end{tabular} & \cite{islam2021exploring}     & SN Computer Science                                                                                                     \\ \hline
2021          & \begin{tabular}[c]{@{}l@{}}Evolution of automatic visual description  \\ techniques-a methodological survey\end{tabular}          & \cite{bhowmik2021evolution}   & Multimedia Tools and Applications                                                                                       \\ \hline
2022          & \begin{tabular}[c]{@{}l@{}}A comprehensive review of the video-to-text  \\ problem\end{tabular}                                   & \cite{perez2022comprehensive} & Artificial Intelligence Review                                                                                          \\ \hline
2022          & Query Focused Video Summarization: A Review                                                                                       & \cite{ akhare2023query}       & \begin{tabular}[c]{@{}l@{}}ISAI: International Symposium \\ on Artificial Intelligence\end{tabular}                     \\ \hline
2022          & \begin{tabular}[c]{@{}l@{}}A study on video semantics; overview, \\ challenges, and applications\end{tabular}                     & \cite{patel2022study}         & Multimedia Tools and Applications                                                                                       \\ \hline
2023          & A Survey on Temporal   Sentence Grounding in Videos                                                                              & \cite{lan2023survey}          & \begin{tabular}[c]{@{}l@{}}ACM \end{tabular} \\ \hline
2023          & A Survey on Video Moment   Localization                                                                                           & \cite{liu2023survey}          & ACM Computing Surveys                                                                                                   \\  \hline
2023          & A Review of Deep Learning for Video Captioning                                                                                          & \cite{abdar2023review}          & arXiv Pre-print                                                                                                  \\ \hline
\end{tabular}
\end{table}

\subsection{Existing Surveys}
The topic of NLP, particularly the generation of captions for images and videos, has attracted much attention in recent years, mainly between 2018 and 2023. This period has witnessed a surge in summarizing the efforts being made in the field. After a thorough survey of literature we found out that most of the survey papers related to caption generation either focus on the general description of the video \cite{singh2020comprehensive, islam2021exploring, bhowmik2021evolution, perez2022comprehensive, aafaq2019video} or video sentence grounding under summarization task \cite{akhare2023query, lan2023survey}. There is another category of surveys that focuses on the semantics of event placement in videos \cite{patel2022study, liu2023survey}. To this end, we find none of the survey focusing mainly on the dense video captioning (DVC) task for all three main attributes in the pipeline of DVC, i.e., \textit{(1)} video feature extraction, \textit{(2) }temporal event localization, and\textit{ (3)} dense caption generation. Table \ref{tab:survey-table} lists the recent surveys in the field of video caption generation, together with the year of publication and the journal/conference where the study was published.

\subsection{Research Methodology}
\begin{figure}[H]
    \centering
    \includegraphics[width=380px]{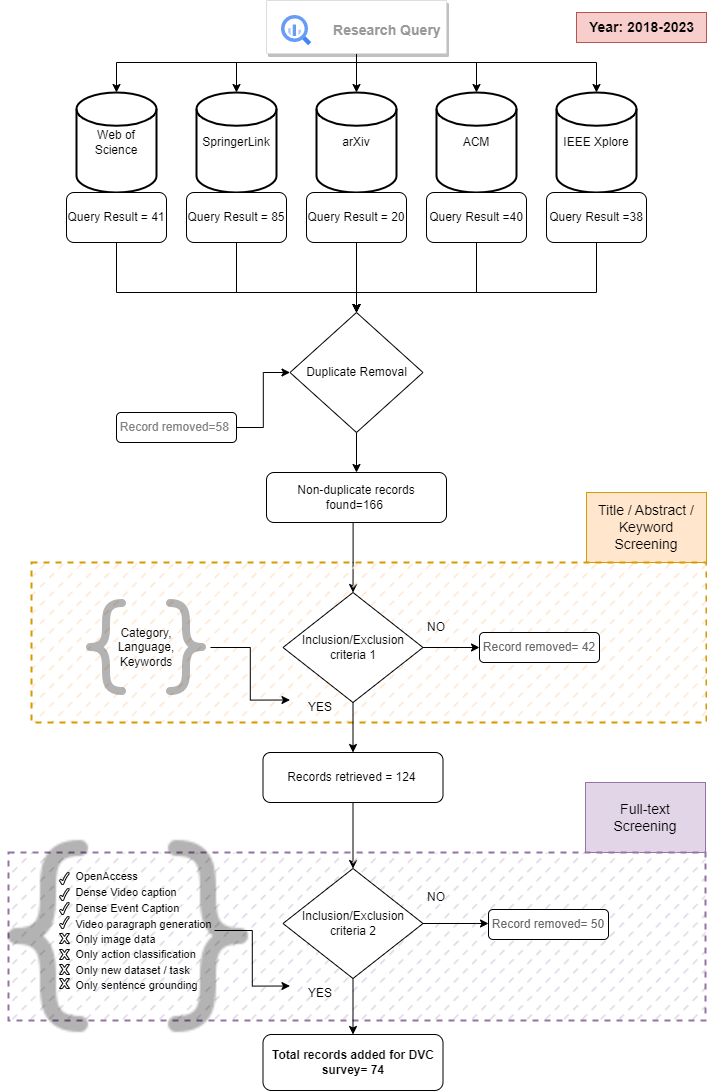}
    \caption{\textit{Flow of Selection Criteria:} A two-stage inclusion/exclusion process is used to match research that fits within the definition of "dense video captioning" techniques. In the first screening stage, category, language, and keywords are considered, while in the full-text screening stage, a more specified selection is made.}
    \label{fig:flow of selection criteria}
\end{figure}

This section presents details of the research methodology adopted for this review (fig. \ref{fig:flow of selection criteria}). We explored 5 databases for Dense Video Captioning review, namely Web of Science, IEEE Xplore, ACM, SpringerLink, and arXiv. The reason for using these databases for research is their prevalent use in the field of machine learning and artificial intelligence. 

Apart from specific research queries based on our interests and relevant databases, there are certain inclusion and exclusion criteria followed for each task, which are discussed below. 

\subsubsection{Inclusion/Exclusion criteria for Dense Video Captioning} Here is how the research methodology is employed to gather insightful data included in the survey. Apart from relevance to the field, studies included in the survey should match the following three criteria: 

\textbf{Date:} Studies from early 2018 till October 2023 are included in this review.

\textbf{Category:} Research articles/journal articles/conference submissions/book chapters and review articles falling in the category of Machine Learning, Artificial Intelligence, Computer Vision, Image Processing, and Pattern Recognition are included in the review. 

\textbf{Language:} Research is restricted to English Language.

\begin{figure}[!ht]
    \centering
    \includegraphics[width=400px]{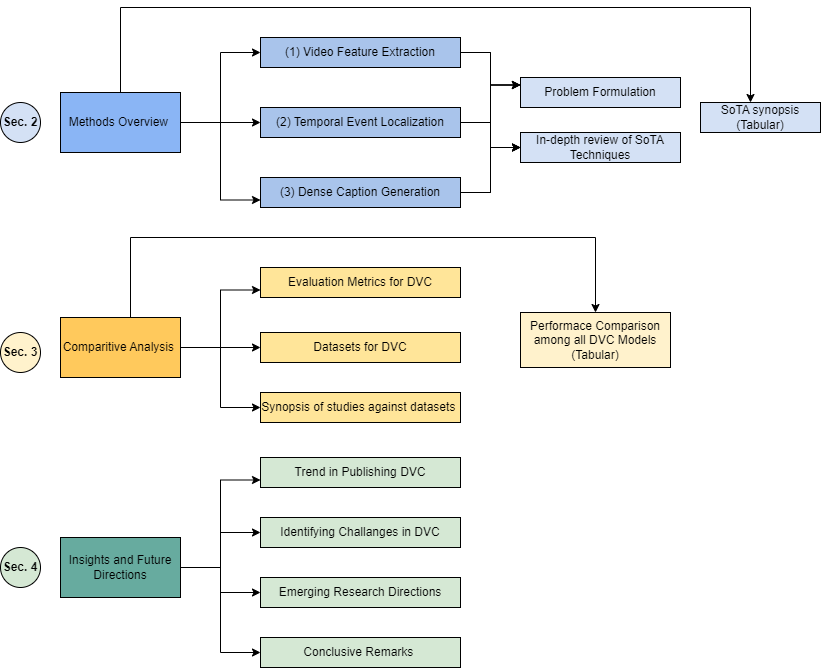}
    \caption{Anatomy of the Review. We present a methodological review of the techniques used for DVC \textit{(Sec. 2)} along with comparative analysis of all the studies \textit{(Sec. 3)} and finally the outcomes and upcoming endeavors \textit{(Sec. 4)}}
    \label{fig:structure of the review}
\end{figure}

\subsubsection{Inclusion/Exclusion criteria for Datasets} The datasets discussed in this survey have been thoughtfully adopted for the studies in DVC. These datasets adhere to four key criteria to ensure utmost relevance to the topic. By carefully selecting and curating these datasets, we aim to enhance the applicability and significance of our review in the field of video captioning.

\textbf{Date:} Datasets used in the research done from 2018 till October 2023 are included. 

\textbf{Description:} Large-scale datasets narrating the event in natural language sentences with timestamp details targeting dense event captioning are included. Datasets that only contain action categories (sitting, dancing, applying makeup) are not considered part of dense event captioning. Furthermore, datasets that belong to the studies     

\textbf{Language:} Datasets that describe the events in English Language sentences are included.

\textbf{Availability:} Datasets that are available publicly or/and are part of an open dense captioning challenge are included.

The survey is organized as shown in Fig\ref{fig:structure of the review}. Section 2 describes DVC approaches in depth, specifies procedures, and examines state-of-the-art (SoTA) methods, with findings summarized in a table. Section 3 includes a comparison of statistics, benchmark datasets, and research for DVC from 2018 to 2023. Section 4 investigates trends, problems, and possible multidisciplinary links in DVC study.

\section{Methods Overview}
\begin{figure}[H]
  \centering
  \includegraphics[width=300px]{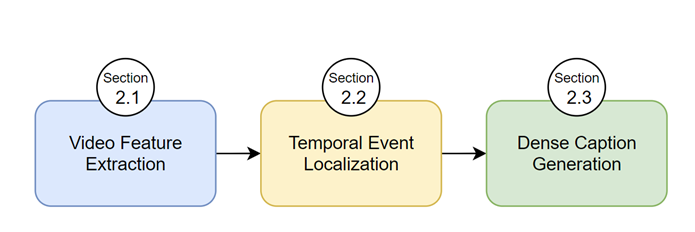}
  \caption{General Workflow of a DVC model: schema followed in the whole review}
    \label{general workflow of a DVC model}
\end{figure}

The general workflow of a DVC model is shown in the fig. \ref{general workflow of a DVC model}. After the pioneering work in DVC by Krishna et al. 
\cite{krishna2017dense}, most of the literature tends to follow the same three-tier architecture. This survey is also organized following the same protocols, i.e., (1) Video Feature Extraction (VFE), (2) Temporal Event Localization (TEL), and (3) Dense Caption Generation (DCG). 

\subsection{Video Feature Extraction}
Usually, features are extracted from input videos and stored in a feature vector, and events are localized using the encoder module.Some captioning methods rely only on the visual-linguistic modality, while others exploit audio features. A few methods do not rely on explicit feature extraction or event proposal generation, so they use pre-extracted features. The following section of the paper delves into various techniques employed to transform raw video data into a more compact and expressive representation Table \ref{tab:video feature table}.

\textbf{Problem formulation:} Given a video $V = \{v_i\}_{i=1}^{|v|}$ with $|v|$ being the number of frames inside video $V$, the objective of video feature extraction is to extract a set of meaningful and informative features that represent the visual content of the video. The extracted features can be denoted as $F = \{f_i\}_{i=1}^{|F|}$, where $|F|$ is the number of features extracted. Each feature $f_i$ is a high-dimensional vector that captures the relevant information about a specific video aspect, such as color, texture, shape, motion, or semantic content.

\textbf{C3D} \cite{tran2015learning} is famous for its widespread use for video analysis, specifically for extracting spatiotemporal features from video data and temporal jittering for data augmentation.  C3D achieved state-of-the-art performance on several video classification benchmarks, demonstrating the effectiveness of spatiotemporal feature learning using 3D CNNs \cite{rahman2019watch, yamazaki2022vlcap, zhang2022unifying, chen2021towards, mun2019streamlined, iashin2020multi, wang2023learning, wang2018bidirectional}. A few recent DVC methods also work with pre-trained C3D or 3D-CNN features \cite{ zhu2022end, li2023generating, li2023time} 

\textbf{VGGish} \cite{meyer2017efficient}
 is an audio feature extraction model based on the VGG architecture, introduced by researchers at Google in 2017 \cite{meyer2017efficient}. It processes raw audio waveforms and transforms them into compact embeddings through a series of steps, including pre-processing, spectrogram computation, normalization, patch creation, and feeding into the VGGish convolutional neural network. These embeddings can then be used for various audio-related tasks such as sound classification, similarity search, or content-based retrieval. Many multi-modal dense video captioning techniques make use of VGGish architecture for audio feature retrieval \cite{iashin2020better, palivela2023dense, chang2022event, han2023lightweight, iashin2020multi, mittal2022savchoi, wang2022semantic, song2020team, chen2018ruc+}. 

 \begin{center}
\begin{table}[h]
\caption{\textbf{Video Feature Extraction Techniques}. We use the following abbreviations. Action Recognition \textbf{(AR)}, Video Analysis \textbf{(VA)}, Audio Classification \textbf{(AC)}, Content-base Audio Retrieval \textbf{(CAR)}, Image Classification \textbf{(IC)}, Object Detection \textbf{(OD)}, Image Processing \textbf{(IP)}, Transfer Learning \textbf{(TL)}}
\label{tab:video feature table}
\begin{tabular}{llllll}
\hline
\textbf{Model} & \textbf{Ref.} & \textbf{Architecture}                                            & \textbf{Model Input}                                                  & \textbf{Pretrained on}                                                             & \textbf{Application}                                                                                             \\ \hline
C3D       & \cite{tran2015learning}         & 3D-CNN                                                           & Video frames                                                    & Sports-1M                                                                & \begin{tabular}[c]{@{}l@{}}AR, VA\end{tabular}                           \\ \hline
I3D        & \cite{carreira2017quo}        & \begin{tabular}[c]{@{}l@{}}3D ConvNet\end{tabular}        & Video frames                                                    & \begin{tabular}[c]{@{}l@{}}ImageNet + \\ Kinetics\end{tabular}          & \begin{tabular}[c]{@{}l@{}}AR, VA\end{tabular}                                                       \\ \hline
VGGish    & \cite{carreira2017quo}         & CNN                                                              & Audio spectrograms                                              & YouTube-100M                                                              & \begin{tabular}[c]{@{}l@{}}AC, CAR\end{tabular}                                   \\ \hline
ResNet    & \cite{he2016deep}         & CNN                                                              & Image frames                                                    & ImageNet                                                                & \begin{tabular}[c]{@{}l@{}}IC, OD, IU\end{tabular}                             \\ \hline
CLIP      & \cite{radford2021learning}         & \begin{tabular}[c]{@{}l@{}}VL model \end{tabular} & \begin{tabular}[c]{@{}l@{}}Image and text \\ pairs\end{tabular} & \begin{tabular}[c]{@{}l@{}}ImageNet + \\ Text corpus\end{tabular} & \begin{tabular}[c]{@{}l@{}}VQA, IC\end{tabular}                                                                  \\ \hline
EfficientNet   & \cite{tan2019efficientnet}    & CNN                                                              & Image frames                                                    & ImageNet                                                                & \begin{tabular}[c]{@{}l@{}}IC, OD, TL\end{tabular}      \\ \hline
\end{tabular}
\end{table}
\end{center}
 
 \textbf{I3D}, or Inflated 3D ConvNets, is an edge-based video feature extraction model introduced by DeepMind researchers in 2017 \cite{carreira2017quo}. By extending 2D convolutional neural networks (CNNs) to 3D, I3D captures spatial and temporal information in videos. The model inflates filters and pool kernels of pre-trained 2D CNNs (e.g., Inception-v1 or ResNet) into 3D, enabling direct learning of spatiotemporal features from video data. I3D achieves exceptional performance in various video classification and action recognition tasks \cite{ iashin2020better, chen2019activitynet, estevam2021dense, zhang2022fine, wang2023learning, han2023lightweight, iashin2020multi, mittal2022savchoi, wang2022semantic, song2020team,chen2018ruc+}, making it a powerful tool for video analysis and hence dense video captioning.

 \textbf{GloVe} stands for Global Vectors for Word Representation. It is a model for learning word embeddings and dense vector representations of words. It starts by constructing a co-occurrence matrix and uses a weighted least squares optimization objective to learn embeddings. The objective function $J$ aims to minimize the squared difference between the dot product of two-word vectors and the logarithm of their co-occurrence count. The equation is as follows:

\[
J = \sum_{i,j=1}^{V} f(X_{ij}) \left( (w_i^T \ast w_j + b_i + b_j - \log(X_{ij}))^2 \right)
\]

The model is trained using optimization techniques like stochastic gradient descent (SGD) to minimize the objective function. After training, GloVe outputs word embeddings that capture semantic meaning and relationships between words, making them suitable for various natural language processing tasks and performing better than word2Vec.

\textbf{ResNet} \cite{he2016deep}, short for Residual Networks, is a type of convolutional neural network that uses "skip connections" or "shortcut connections" to mitigate the problem of vanishing gradients, which makes it possible to train intense networks. Video feature extraction network proposed by \cite{lee2021dvc} uses C3D and ResNet to extract features from video data. The adversarial interface also uses ResNet and R3D to perform feature extraction \cite{kanani2021global}. The ResNet200 is a variant of ResNet that includes 200 layers. It is designed to learn rich feature representations from image data and adopted for video feature extraction \cite{prudviraj2022aap, chen2019activitynet}.

\textbf{CLIP} serves as an epitome of "zero-shot" learning, i.e., it can get adopted for the tasks it was not explicitly trained on, making it a versatile model for a wide variety of language and image tasks \cite{radford2021learning}. CLIP is introduced in a contrastive manner; it measures the contrastive loss as follows:
\[
L = - \frac{1}{N} \sum \log \left( \frac{\exp(s(i,i'))}{\sum \exp(s(i,j))} \right)
\]
where $N$ is the batch size and $s(i,j)$ is the scaled dot-product similarity between image $i$ and text $j$. This loss function encourages the model to produce image and text representations that are similar (in terms of dot product) for positive pairs (i.e., corresponding images and texts) and different for negative pairs (i.e., non-corresponding images and texts). By minimizing this loss, the CLIP model learns to align the image and text representations in a shared embedding space, which enables it to perform a wide range of vision and language tasks. State-of-the-art Vid2Seq \cite{yang2023vid2seq}, VLCAP \cite{yamazaki2022vlcap}, and VLTinT \cite{yamazaki2022vltint} use CLIP for linguistic feature extraction.

\textbf{EfficientNet B7} is one of the model variants in the EfficientNet family of convolutional neural networks (CNNs) designed for image recognition and feature extraction. Introduced by researchers at Google AI, the EfficientNet architecture is built upon the idea of compound scaling. It involves scaling the model depth, width, and input resolution simultaneously to achieve better performance while maintaining efficiency. The depth, width, and resolution of the network are determined as follows:
\[
\text{{depth}} = \alpha^\varphi
\]
\[
\text{{width}} = \beta^\varphi
\]
\[
\text{{resolution}} = \gamma^\varphi
\]
where $\alpha$, $\beta$, and $\gamma$ are constants that determine how much the depth, width, and resolution should be scaled, and $\varphi$ is a user-defined scaling factor that controls the overall scaling of the model.
The EfficientNet family includes a series of models (B0 to B7) with increasing complexity and capacity, where EfficientNet B7 is the largest and most powerful model among them for video feature extraction \cite{madake2022dense}.

\subsection{Temporal Event Localization}
Selecting the start and end time of an event inside a complex video to address the semantics of the video along with its temporal dimensions is known as \textit{Temporal Event Localization}. The spatiotemporal component of videos should be looked at during event identification, either at a minimal level with the right features or at a greater level with algorithms that can leverage temporal sensitivity. 

\begin{figure}[H]
  \centering
\includegraphics[width=380px]{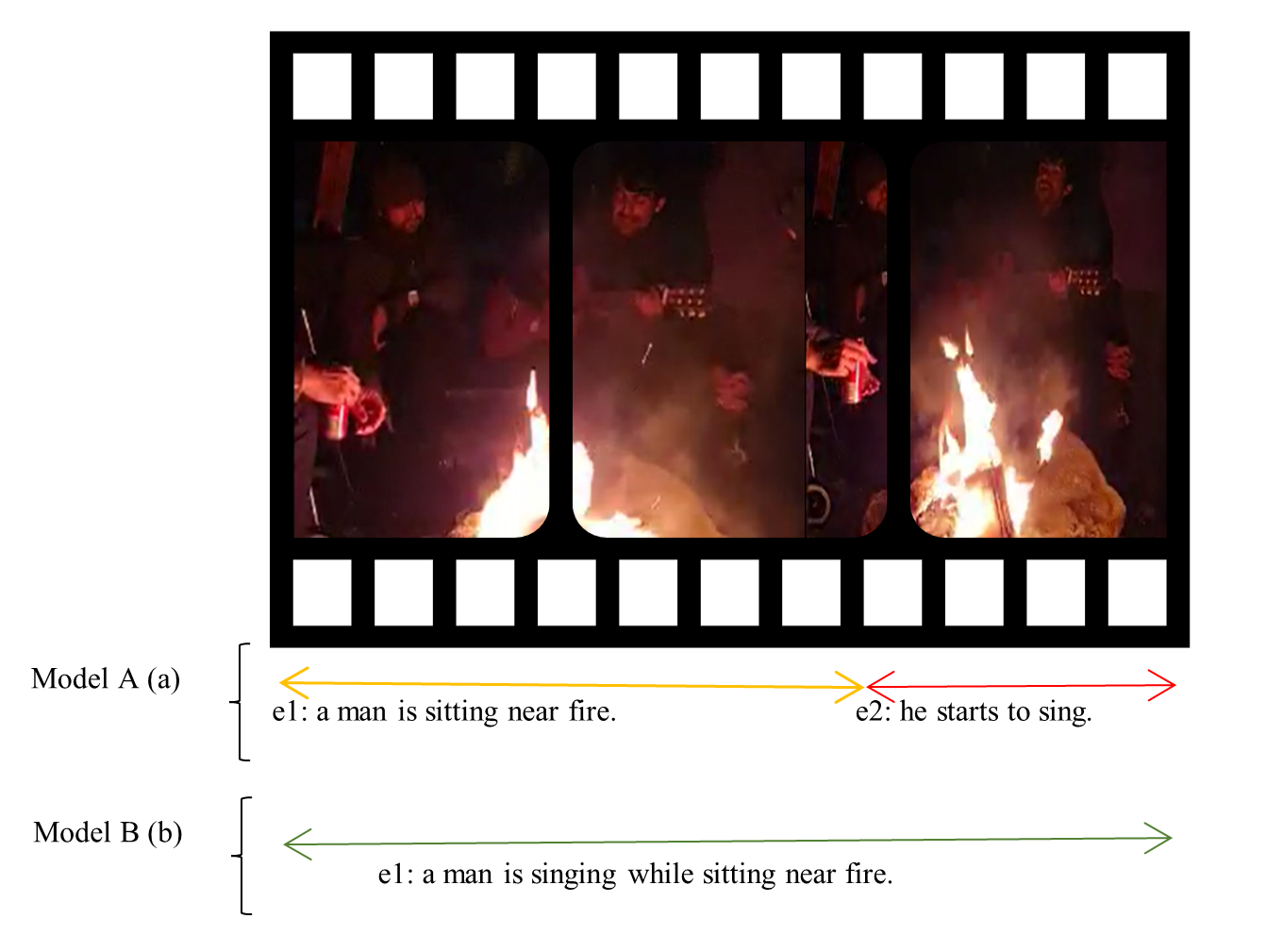}
  \caption{An example of events proposed by two separate models}
  \label{example of event proposal}
\end{figure}

In literature, the definition of an event is contradictory; it can be a single action or a set of actions represented as a single event. For example, "a man is sitting near the fire" and "he (the man) starts to sing" can be proposed as two separate events by one proposal module Figure \ref{example of event proposal}(a). Meanwhile, another module may capture it as a single event, saying, "a man is singing while sitting near the fire" Figure \ref{example of event proposal}(b). This problem occurs due to (1) the undefined time limit of an event since the time limit cannot be decided in advance, (2) the set of actions defined in pre-training models; some models tend to delve into deeper details of the scene by adding more verbs and adverbs to capture minute video details while ignoring the sentence length, (3) the modalities (Audio, Video, Text) involved in feature extraction also impact the proposed event. Moreover, other coarse-grained information in the training dataset also influences the proposed events and, consequently, the generated captions. Thus, models that claim to handle dense video captioning need to have a robust proposal module before generating captions.

\textbf{Problem formulation} Providing a Video, $V=\{v_i\}_{i=1}^{|v|}$ with $|v|$ being the number of frames inside video $V$, the objective of event localization is to precisely separate each unique event $e$ in the video such that $e_i=\{e_i^{start}, e_i^{end}\}_{i=1}^{|E|}$, where $|E|$ is the number of events $e$ and $e_i^{start}$ and $e_i^{end}$ depict the start and end of event $e_i$ respectively. The keyframes for the detected event $e_i$ can be presented as $v_e=\{v_i | e^{start} \leq i \leq e^{end}\}$.

\begin{figure}[H]
  \centering
  \includegraphics[width=380px]{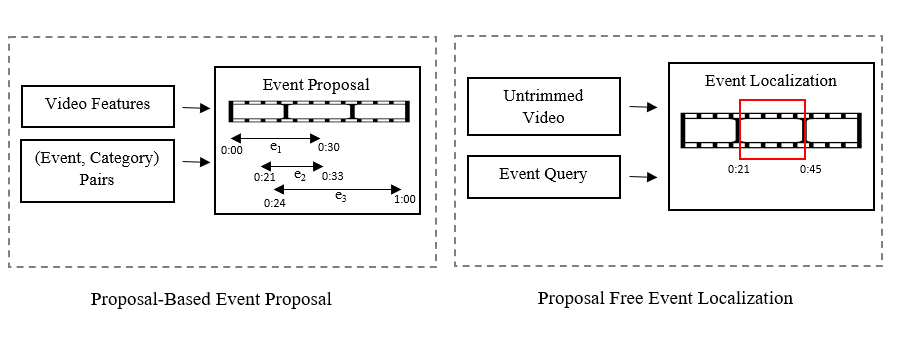}
  \caption{Side by side simple block diagram of two Temporal Event Localization (TEL) techniques}
  \label{fig:Difference between proposalfree and proposalbased}
\end{figure}

Temporal Event Localization (aka TAPG-Temporal Action Proposal Generation) module requires extracted features from a video as input and outputs localized events with timestamp details. Secondly, it infers the action category of that event. An ideal localization module should include a small number of proposals with high recall value and high Temporal Intersection over Union (tIoU) value. Research in TEL can be broadly divided into two main categories, \textit{(1) Proposal-based methods}, and \textit{(2) Proposal-Free methods}. The general workflow diagram of proposal-based and proposal-free TEL is shown in Figure \ref{fig:Difference between proposalfree and proposalbased}. We discuss the methods based on their functionalities/work pipeline and further categorize each way into its sub-categories. Under all sub-categories, we discuss state-of-the-art methods. 

\subsubsection{Proposal-based Methods}
Proposal-based techniques use an algorithm or module to offer potential event proposals in videos, which are then scrutinized and improved to produce precise event localization.
 
\textbf{Sliding Window:} Traditionally, temporal events were localized as anchors using a sliding window, and the proposals with the highest score were selected as proposed candidates. Selecting a window size is tricky as it impacts action fragments \cite{yuan2016temporal}. One of the primary efforts towards temporal localization of events for long videos started back in 2016 with the \textbf{ Deep Action Proposals (DAPs) } architecture by Escorcia et al. \cite{escorcia2016daps}. It leveraged LSTM for encoding video content and forecasting suggestions covered by the sliding window. In 2017, Buch et al. \cite{buch2017sst} proposed \textbf{Single-Stream Temporal (SST)} action proposals mechanism that processes whole video in a single stream and does not use memory batches. Contradictory to DAP, SST produced non-overlapping sliding windows for event localization, achieving higher average recall. SST uses k different offset values to densely produce proposals in one direction. \textbf{Temporal Actionness Grouping (TAG)} \cite{zhao2017temporal} gathers the temporal structure of activities and exploits that knowledge to distinguish between complete and incomplete events. TAG treats proposal generation as a regression problem. The sliding window starts at 0.3 seconds and jumps at 0.4 each time with 20 exponential scales. \cite{shou2016temporal} also used sliding window fashion for proposal, classification, and localization of events in video data. \textbf{Proposal-level Actionness Trustworthiness Estimator (PATE) }\cite{Gao_2018_ECCV} exploits both, the sliding window manner with CTAP containing CNN boundary awareness for temporal action proposal generation task. \cite{chen2018ruc+} applied a heuristic sliding window and proposal ranking based on the highest tIoU value. \textbf{Event Sequence Generation Network (ESGN) }\cite{mun2019streamlined} uses SST along with PtrNet \cite{vinyals2015pointer} to select highly correlated events in order. The idea of event sequence generation using ESGN has been adopted in other DVC techniques as well \cite{chen2019activitynet,yu2021dense}. A few other techniques such \textbf{TL-NMS} \cite{wang2020event}, \textbf{HEAD} \cite{yu2021dense}, \textbf{PDVC} \cite{ wang2021end}, \textbf{Prompt Caption Network (PCNet)} \cite{li2023generating} and \textbf{Key Point Positioning}\cite{zhang2022fine} also used SST with ESGN to localize events in the videos temporally. In \textbf{Bi-SST framework}, such as \cite{iashin2020multi}, events are proposed during forward and backward passes on a video sequence.

\textbf{\textit{Boundary Aware Networks:}} Boundary-aware mechanisms directly target event boundaries, i.e., event start and end information. These methods use object bounds and coarse-to-fine strategy to predict video events by comparing extracted features from the already defined representation of the target event. \cite{Lin_2018_ECCV} introduced \textbf{Boundary-Sensitive Network (BSN)} that adopted a ‘bottom-up’ approach to locate events and work precisely with them using confidence score. Another end-to-end proposal generation mechanism, \textbf{Boundary-Matching Network (BMN)}\cite{lin2019bmn}, addressed the problem of evaluating the confidence score for each densely distributed proposal without an anchor mechanism. BMN introduced a ‘Top-down’ approach and simplified the work pipeline with the combined idea of \cite{lin2017single} and \cite {Lin_2018_ECCV}. 

Reinforcement learning techniques are well suited for extracting events in boundary-aware scenarios. Such as, a variant of BAN, called \textbf{MABAN} \cite{9451629} uses reinforcement learning to extract contextual semantic knowledge of video to generate action proposals. The temporal regression module helps to deduce temporal event boundaries in the MABAN model. Likewise, \textbf{Dense Boundary Generator (DBG)} \cite{lin2020fast} uses an action-aware regression module to infer high-level actions after low-level boundary classification. \cite{wang2020dense} also used DBC along with EGGN and proposed an RNN-based hierarchical semantic aware modal for better event-level representation. \textbf{State-aware LSTM encoders} use boundary information to extract discontinuity between proposed and GT frames to generate proposals \cite{baraldi2017hierarchical}. 

\cite{suin2020efficient} also devised a frame selection network that used guided reinforcement learning for efficient DVC. Given a Video, $V={v_i}{i=1}^{|v|}$ with $|v|$ number of frames per frame and $Y$ sentences per frame $i$, such that $Y={y_i|i\in[1,n]}$, DepNet \cite{bao2021dense} uses boundary matching mechanism to embed learned features with start $s$ and end $e$ of the proposal:
\[
\widetilde{f}^{v{st}}=BM({v_i}{i=s}^e)
\]
After generating possible moment proposals, the visual semantic proposal generator extracts positional features and passes them on to the aggregation module. This propagation module then computes temporal moment score $\widetilde{M}^{k{st}}$ to match temporal sentence score $\widetilde{p}^{k_{st}}$. A few anchor-free mechanisms \cite {yang2020revisiting} also exploits neighboring boundary knowledge for temporal event proposal generation task. 

\textbf{\textit{Context gating mechanisms:}}  Context-aware localization systems employ signals to localize events according to the user's context. They create more human-like captions by utilizing visual, auditory, or cognitive embeddings and comprehending typical sequences such as "knead the dough" followed by "mix water in dough" \cite{ji-etal-2021-hierarchical} and were first introduced in 2018. \textbf{Bi-directional Single-Stream Temporal (Bi-SST) mechanism} \cite{8578849}is a variational model to the existing SST. Bi-directional SST employs context gating to capture both past and future contexts since single-stream models cannot accommodate the context of future events. 
Wang et al. also incorporated an attention fusion mechanism to avoid recurring proposals. However, these methods are usually detected by classification-based methods. Confidence scores of each proposal from both passes are then fused using simple multiplication: 
\[
C_p = c_i \cdot c_{i'} = \frac{1}{N}
\]
where $\vec{c}_i$ represents the forward pass confidence value, $c_i$ represents the backward pass value, and $N$ represents the total number of proposals. BiLSTM \cite{madake2022dense} uses previous and future context information for temporal event localization. \\\textbf{ProcNets Procedure Segmentation Networks} consist of context-aware frame-wise encoding module and segment proposal module \cite{zhou2018towards}. After localizing each event's start and end point, another module called 'sequential prediction' gives the final verdict of proposed segments using LSTM. The network works in a self-supervised manner for long videos. ProcNets followed human learning harmony instead of using subtitles and pre-trained knowledge. They define the generation of proposals as a classification problem. \cite{zhou2018towards} also introduced a dataset with temporally localized procedures to do the task and called it YouCook2. Inspired by ProcNets, Zhou et al. \cite{zhou2018end} used an anchor-free mechanism to exploit event proposal detection. \\ \textbf{Dense Procedure Captioning (DPC)} \cite{shi2019dense}, unlike ProcNets, generates a feature matrix using transcript embeddings with video frames. Transcript embeddings $e \in \mathbb{R}^2$ and video embeddings $e \in \mathbb{R}^{T \times d}$ are encoded using BERT-L and ImageNet pre-trained ResNet32, respectively, and then passed through a transformer layer. The resulting embeddings are fed into the context-aware model. \textbf{Graph Convolutional Network, G-TAD} \cite{xu2020g} incorporates multi-level semantic concepts into video features and casts temporal action localization as a graph localization problem. It aggregates the context of every cluster to understand its features, constantly updating a network's connections. The GCNeXt module is a context encoder that learns the semantic and temporal information from the input snippet features. SGAlign signifies the sub-graphs as feature vectors. All the snippets of videos are graph nodes, and the snippet-to-snippet relation is depicted as graph edges. Lastly, the localization module works in a 'score and rank' manner to find proposals. For a given Video, $V={v_i}_{i=1}^{|v|}$ and node and edge sets $\varepsilon=\varepsilon_t \cup \varepsilon_s$. G-TAD creates a graph $G={V,E}$ to exploit multi-level semantic context. \textbf{Stacked Multi-modal Attention Architecture SMAN }\cite{9351972} applied a stacked architecture based on reinforcement learning to refine proposed events gradually. The model integrates visual and textual information as context before further improving proposed events via coarse-to-fine training.

\textbf{\textit{Probing For Event Retrieval:}} Natural language-based query contains domain knowledge and cues for better sentence structure that further helps better understand events in long videos. Video-query-based methods are also used to localize events in videos using probs. These methods use natural language queries to see 'what's happening' inside the video. For example, \textbf{Semantic Activity Proposal (SAP)} \cite{Chen2019SemanticPF} uses semantic characteristics of research query to extract events from videos. Unlike the 'propose-classify' manner, SAP uses reranking for coarse-to-fine proposal generation. Shin et al. \cite{shin2022learning} exploits the usefulness of the human cognitive system and developed a Cross-Modal LSTM called CM-LSTM. For temporal event localization, the model relies on the query-based contextual semantics of video along with "\textbf{TACI, a two-stream attentive cross-modal interaction network}."  
\textbf{SAVCHOI} \cite{mittal2022savchoi} also used probing-based anchors to exploit human-object interaction detection in surveillance videos guided by DVC. 

\textbf{\textit{Temporal Networks:}} Temporal networks are time-varying networks that are inherently dynamic in nature and help to extract persistent and recurring patterns in videos. \textbf{Temporal Segment Network (TSN)} uses segment-based modeling to exploit the temporal relationships between video frames to capture events. Some methods work with channel-wise information in video frames, such as \textbf{Channel-wise Temporal Attention Network (CTAN)} \cite{lei2019channel} uses a temporal information aggregation module to exploit temporal relationships in the video. Similarly, \cite{wang2018temporal} also works on segment-based sampling and aggregation modules. Another segment-based mechanism, \textbf{ASTN} \cite{sun2019atsn}, uses human-like visual attention to capture constructive features. Shou et al. proposed a 3D ConvNets consisting of proposal, classification, and localization in which the proposal section helps identify the segments in the video. The deep learning-based attention mechanism in \cite{yang2022sta} called \textbf{STA-TSN} uses the spatiotemporal key feature. Their architecture uses a multi-scale feature enhancement strategy and pyramid pooling to help TSN exploit temporal dynamic features. A transformer-inspired joint learning modal, BIVT \cite{nishimura2022recipe} also uses TSN, Gumbel SoftMax re-sampling, and MIL-NCE. \textbf{Temporal Event Proposal TEP} \cite{li2018jointly} classifies temporal event proposal as a regression problem and works with object boundary localization. TEP is trained to get a lower loss value for multi-task loss. Temporal coordinate regression loss with trade-off parameters $\alpha L_{TCR}$ , descriptive regression loss with tradeoff loss $\beta L_{DES}$ and event loss $L_e$:  
\[
L_{TEP}\ =\ L_e\ +\ \alpha L_{TCR}+\alpha L_{DES}
\]

\textbf{\textit{Mix-model architectures:}} Events are not merely localized using the visual and textual information of the videos. Sometimes, videos have audible cues that help in understanding the video's semantics and localizing events. Mix-modal architecture is the method that uses multi-modality for generating proposals, such as \cite{iashin2020multi}. Secondly, mixed-modal architecture also states the method that neither works in a top-down nor a bottom-up fashion. This category of the TEL jointly localizes while describing the video events \cite{aafaq2021cross, zhu2022end, zhou2018end, xu2019joint, li2018jointly}. \\ \textbf{Bi-modal multi-headed proposal generator} \cite{iashin2020better} make use of a transformer-based encoder-decoder network. The proposal generator comprises a proposal head, common pool, and select and sort mechanism. It exploits audio and visual features to generate proposals and confidence scores. The encoder block then re-represents features for better proposal generation. Several other methods tweak the transformer block with attention, such as \textbf{APP-MIT}\cite{prudviraj2022aap}. \textbf{JEDDi-Net} \cite{xu2019joint} first uses a simple region Convolutional 3D Network (R-C3D) model and max-pooling filter to remove the spatial dimension from the feature vector. Next, for each temporal event $e_i$, the model predicts the center value and length of the proposed event such that $\left\{e_c{,\ e}_l\right\}$. Apart from that, a few models also use a pre-defined set of events. For instance, \textbf{Semantic concept classification (SCC)} \cite{wang2022semantic} used predefined/ pre-localized events from \cite{yu2021dense}.

 \subsubsection{Proposal-free methods:}
 With proposal-free video captioning, sentences are generated or grounded from the video frames without requiring a preliminary phase of event proposal modules such as start-end temporal pairs. Instead, captions are rendered using random text corpora \cite{nag2022proposal,nam2021zero}. Proposal-free methods are computationally reliable because they do not suffer from the cost of generating explicit events \cite{Zhao_2021_CVPR,wang2021end} and limited classes \cite{lin2017single}. Natural language query-driven, \textbf{Temporal Video Grounding (TVG)} is one of the successful techniques \cite{Rodriguez_2020_WACV,Zhao_2021_CVPR, zhang2020does, chen2018temporally} for identifying actions and events in long videos in a proposal-free way. 
 
Some of the weakly supervised and unsupervised dense captioning techniques prefer to use the proposal-free methods. Pioneer work in hand-driven heuristics regarding proposal-free methods for TEL is \textbf{Moment Context Network (MCN}) \cite{Hendricks_2017_ICCV}. Gao et al. also worked similarly and proposed \textbf{Cross-modal Temporal Regression Localizer (CTRL)} \cite{gao2017tall}. These techniques use shared embeddings of video features and language knowledge for localizing events. The Famous THUMOS and DiDeMo datasets of clip-expression pairs are associated with MCN and CTRL, respectively. Proposal-free methods are best suited for scenarios without much-annotated data for pre-training, such as medical image analysis for microscopic videos.

\subsection{Dense Caption Generation (DCG)}
\begin{figure}[h]
    \centering
    \includegraphics[width=400px]{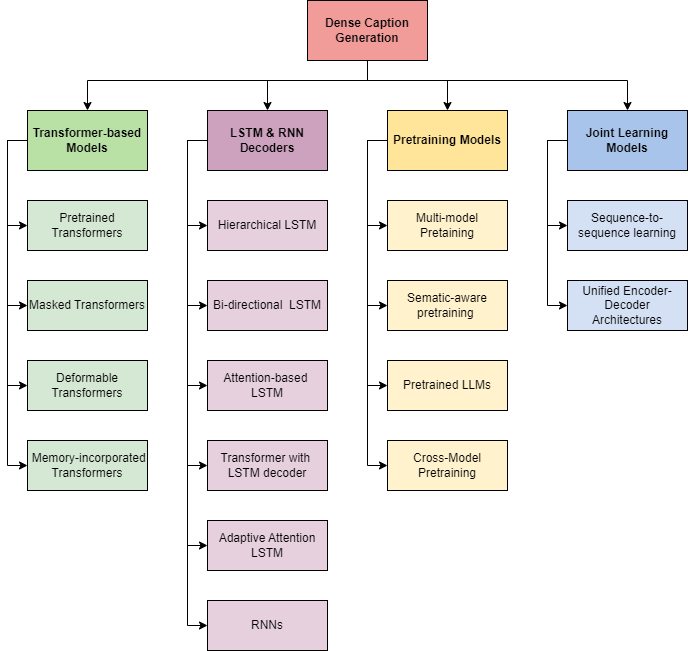}
    \caption{Taxonomy of DCG Techniques(2018-2023)}
    \label{fig:DCG flowchart}
\end{figure}

After the ‘detect’ part, the next step of DVC is to ‘describe’ the events in natural language. Some fundamental DCG methods are listed in fig. \ref{fig:DCG flowchart}. A single technique or an amalgamation of two or three strategies may be employed for a dense video captioning challenge. 
\\
\textbf{Problem formulation:} Given a video $V = \{v_i\}_{i=1}^{|v|}$ with |v| being the number of frames inside video \textit{V}, the objective of Dense Caption Generation is to generate a sequence of natural language descriptions that capture the visual content of the video at different time steps. Each description or caption $c_i$ corresponds to a segment of the video between frame ${t_i}^{start}$ and frame ${t_i}^{end}$ and should be able to convey the relevant information about the visual objects, actions, and events that occur in that segment.

\subsubsection{Transformers for DCG}
Transformers are fundamentally deep neural networks carefully designed to capture complex connections between variables inside sequences. In this survey, 19 out of 74 studies have used a dedicated transformer block in the captioning part of their DVC model. A few others have incorporated it as a part of mixed-modal architecture. After the success of the end-to-end \textit{masked Transformer} \cite{zhou2018end}, other studies follow the trend, such as \cite{yu2021accelerated, suin2020efficient, dang2021dense}. The second class of transformers models used in the DCG part are \textit{pre-trained transformers}, such as Vanilla Transformer \cite{estevam2021dense, chen2018ruc+}. KeyBERT SBERT \cite{palivela2023dense}, BERT \cite{yan2021dvcflow}, pretrained PDVC \cite{zhang2022fine}. GloVe pre-trained word embedding is also used in architectures such as \cite{han2023lightweight, nishimura2022recipe}.\textit{Deformable transformers}  modify the regular transformers by introducing deformable attention-block to adjust their attention patterns dynamically. They also prevail in the recent dense caption generation models such as \cite{wang2021end, choi2022parallel, scherer2022semantic}. Another form of Transformer being used in the studies under consideration is \textit{memory-incorporated Transformer} \cite{prudviraj2022aap}, \textit{recurrent transformer} \cite{li2023time}, and \textit{Transformer in Transformer (TinT)} \cite{yamazaki2022vltint}.

This section discusses the basic functionalities of all the dense caption generation techniques that use transformer blocks. Firstly, Zhou et al. \cite{zhou2018end} introduced the concept of a Transformer for DVC by replacing the LSTM decoder with the transformer backbone. Later, Shizhe Chen et al. used a caption generator procedure that involves training a caption model based on ground truth event proposals using cross-entropy loss. The model is then fine-tuned with self-critical reinforcement algorithms using rewards from METEOR and CIDEr \cite{chen2019activitynet}. Iashin et al. \cite{iashin2020better} developed a Transformer-based bimodal transformer (BMT) for caption generation based on audio-visual cues, creating captions word-by-word until an end token is achieved . Huang et al. created a transformer-based MPDVC for complicated caption with graphical and linguistic inputs, utilizing Masked Sequence-to-Sequence (MASS) pretraining and validation on two datasets with segment-level captions \cite{huang2020multimodal}.

The Caption Generation Transformer \textbf{VSJM-Net} by Nayyer et al. decodes input embeddings into plain language descriptions using a transformer design. Sub-task modules provide dense captions. Encoder-decoder multi-headed attention and a Fully Connected Feed-forward Network (FFN) generate the next word with the maximum probability \cite{aafaq2022dense}. To group the advantages of LSTM and Transformer, Liang et al. developed a cascade of LSTM and transformer-based decoders, \textbf{REASONER}. An initial caption is created using a transformer-based decoder, and it is improved by a series of LSTM-based decoders that employ abductive reasoning. The model selects the best caption from a pool of candidate captions at each reasoning step based on a scoring function that considers visual and linguistic cues to provide a final predicted caption \cite{liang2022visual}. 

Instead of using heavy bi-modal architectures, Han et al. suggested \textbf{CMCR}, a cross-modal decoder, to model the subsequent caption word distribution matching the vocabulary size of the training set. The decoding block will utilize the words created in the previous time step to anticipate the expressions in the current time step \cite{han2023lightweight}. Moreover, they proposed a commonsense reasoning (CR) module to optimize the logic of generated captions.

Recently, Yamazaki et al. \cite{ yamazaki2022vlcap, yamazaki2022vltint} defined intra and inter-event inside the video and proposed two models \textbf{VLCAP} and \textbf{VLTinT} to generate coherent captions for these events. Both models are inspired by unified transformer encoder-decoder architectures and are built on top of MART. VLCAP uses visual-linguistic features to capture the semantically relevant representations. The text transformer in the VLCap decoder uses the sentence, event embeddings, and GRU-like memory to generate captions. The VLTinT, on the other side, uses Hybrid Attention Mechanism (HAM) to capture linguistic features. The models use cross-entropy and contrastive loss to optimize caption generation. 

\textbf{Memory Incorporated Transformers.} The Memory incorporated transformer (MIT) is a transformer-based architecture that employs an external memory module to store information from past time steps. Such transformers enable the model to make more accurate predictions to handle lengthy sequences of inputs. Prudviraj et al. create multi-sentence video descriptions using MIT with multi-head attention, and LSTM-GRU inspired Memory updating block \cite{prudviraj2022aap}. The \textbf{Time-Frequency Memory (TFM)} in \textbf{TFTD} not only retains the information of past and generated captions but also stores the action patterns for coherent caption generation.
\textbf{Deformable Transformer.} Deformable Transformer models in natural language processing use learnable offsets and scales to focus on different feature map positions. Such as, \textbf{PDVC} by Teng Wang et al. generates captions with a decoder and one of three parallel heads \cite{wang2021end}. The augmented representations of event queries are sent simultaneously into the localization head and caption head as part of a similar set prediction task, resulting in the captions. Vanilla LSTM-based Lightweight Captioning head and a Standard Captioning head using FC Layer followed by softmax. This results in a strong relationship between these two subtasks, which optimization strengthens.

\textbf{Pretrained Transformer.} Pretrained transformers are transformer-based language models trained on massive volumes of data to acquire natural language patterns and features. Pretrained transformers are usually employed to exploit the unsupervised learning strategy and reduce the computation costs and such as Valter Estevam et al. used unsupervised semantic information extracted from video frames, trained on ground truth events and statements to improve the model's captioning abilities\cite{estevam2021dense}.

BERT-based language models are also used as pre-trained models for DCG tasks. For example, DVCFlow \cite{yan2021dvcflow} used a BERT-inspired pre-trained language model for video captioning. In DVCFlow, a dual-encoder architecture simulates video information flow and encodes visual and textual data. DVCTM model designed by Palivela et al. \cite{palivela2023dense} also leveraged KeyBERT and SBERT for keyword extraction and IBM NLU for subject modeling. This approach identifies important keywords from video and audio elements for captions.

\begin{singlespace} 
\footnotesize
\tabcolsep=0.1cm
\centering
\begin{longtable}[c]{@{}llll@{}}
\caption{Comprehensive overview of the studies. Each model is horizontally distributed into three further sections, Video Feature Extraction, Temporal Event Localization, and Dense Caption Generation, respectively. We comprehend the primary approach for each of these three sections followed in each study.}
\label{tab: Comprehensive overview of studies}\\
\toprule
\multicolumn{1}{c}{\textbf{Model}} &
  \multicolumn{1}{c}{\textbf{\begin{tabular}[c]{@{}c@{}}(1)\\ Video Feature Extraction (VFE)\end{tabular}}} &
  \multicolumn{1}{c}{\textbf{\begin{tabular}[c]{@{}c@{}}(2)\\ Temporal Event Localization(TEL)\end{tabular}}} &
  \multicolumn{1}{c}{\textbf{\begin{tabular}[c]{@{}c@{}}(3)\\ Dense Caption Generation(DCG)\end{tabular}}} \\* \midrule
\endfirsthead
\multicolumn{4}{c}%
{{\bfseries Table \thetable\ continued from previous page}} \\
\toprule
\multicolumn{1}{c}{\textbf{Model}} &
  \multicolumn{1}{c}{\textbf{\begin{tabular}[c]{@{}c@{}}(1)\\ Video Feature Extraction\end{tabular}}} &
  \multicolumn{1}{c}{\textbf{\begin{tabular}[c]{@{}c@{}}(2)\\ Temporal Event Localization\end{tabular}}} &
  \multicolumn{1}{c}{\textbf{\begin{tabular}[c]{@{}c@{}}(3)\\ Dense Caption Generation\end{tabular}}} \\* \midrule
\endhead
\begin{tabular}[c]{@{}l@{}}DVCTM\\ \cite{palivela2023dense}\end{tabular} &
  \begin{tabular}[c]{@{}l@{}}Visual features with I3D, \\ Audio features with VGGish, \\ Optical flow features using Raft, \\ Speech using SRT, \\ Text embedding\end{tabular} &
Class scores and Uncertainty modeling  &
  \begin{tabular}[c]{@{}l@{}}Keyword extraction using KeyBERT + \\ SBERT and IBM NLU used for \\ topic modeling\end{tabular} \\* \midrule
DVCL \cite{qian2023dense} &
  \begin{tabular}[c]{@{}l@{}}2D temporal differential CNN + \\ DETR for feature encoding\end{tabular} &
  \begin{tabular}[c]{@{}l@{}}DIoU for event matching followed \\ by TIoU for proposal evaluation\end{tabular} &
  LSTM based local attention \\* \midrule
PCNet\cite{li2023generating} &
  Pretrained CNN e.g., C3D &
  Baseline PDVC followed  &
  Baseline PDVC folowed \\* \midrule
ZeroTA \cite{jo2023zero} &
  CLIP for VFE &
  \begin{tabular}[c]{@{}l@{}}Soft moment masking with pairwise\\ temporal loss and joint optimization\end{tabular} &
  \begin{tabular}[c]{@{}l@{}}Pretrained GPT2 accompanied with \\ vision and language loss \\ for   optimization\end{tabular} \\* \midrule
Mpp-Net \cite{wei2023mpp} &
  Does not extract features &
  Hierarchical Temporal Spatial Summary &
  Multi-Perspective Perception Decoder \\* \midrule
VidL \cite{wang2023learning} &
  C3D + TSP + I3D &
  \begin{tabular}[c]{@{}l@{}}Cross-modal interaction + \\ Semantic-aware label assignment\end{tabular} &
  LSTM-based lightweight text generator \\* \midrule
SBS \cite{choi2023step} &
  C3D for VFE &
  \begin{tabular}[c]{@{}l@{}}Temporal event counter and boundary \\ classifier based on CNN for   better EPG\end{tabular} &
  \begin{tabular}[c]{@{}l@{}}LSTM based sequential decoder with \\ contextual encoder\end{tabular} \\* \midrule
\begin{tabular}[c]{@{}l@{}}EAMA-DVC \\ \cite{wei2023enhancing}\end{tabular} &
  Transformer unimodal encoder &
  Used pre-trained BMT \cite{iashin2020better} &
  Multimodal decoder \\* \midrule
\cite{huang2023fusion} &
  I3D + C3D + VGGish &
  Confidence Module for event detection &
  LSTM decoder \\* \midrule
CMCR \cite{han2023lightweight} &
  \begin{tabular}[c]{@{}l@{}}Audio - VGGish   \\ Visual - I3D\\    Textual – Word Embeddings\end{tabular} &
  \begin{tabular}[c]{@{}l@{}}Cross-modal processing (CM) module \\ Cross-modal attention fusion \\ + Event Refactoring algorithm\end{tabular} &
  \begin{tabular}[c]{@{}l@{}}Commonsense Reasoning (CR) module \\ along with GloVe +  heterogeneous \\ prior knowledge and entities’ association \\ reasoning\end{tabular} \\* \midrule
TFTD \cite{li2023time} &
  Pre-trained 2D-CNN and 3D-CNN &
  Frequency modeling and diversity loss &
  \begin{tabular}[c]{@{}l@{}}Word probability with pretrained\\  memory recurrent transformer\end{tabular} \\* \midrule
Vid2Seq \cite{yang2023vid2seq} &
  CLIP for VFE&
  \begin{tabular}[c]{@{}l@{}}Encoder-based event sequence generation \\ using pretraining and event sequence \\ prediction using fine-tuning\end{tabular} &
  \begin{tabular}[c]{@{}l@{}}Language modeling using visual and \\ speech embeddings in decoder block\end{tabular} \\* \midrule
\begin{tabular}[c]{@{}l@{}} AAP-MIT \\ \cite{prudviraj2022aap} \end{tabular} &
  ResNet-200 + BNInception &
  Temporal correlation attention &
  Memory incorporated transformer \\* \midrule
\cite{madake2022dense} &
  EfficientNet B7 network &
  PyScene Detect API and BiLSTM encoder &
  LSTM decoder \\* \midrule
VSJM-Net \cite{aafaq2022dense} &
  Visual-semantic embedding 2D-CNN &
  Used pre-localized events &
  \begin{tabular}[c]{@{}l@{}}Transformer decoder with attention \\ + word embeddings\end{tabular} \\* \midrule
\cite{zhu2022end} &
  Pretrained 3D CNN features &
  \begin{tabular}[c]{@{}l@{}}Event detection as segmentation problem \\ to generate target string using tagging \\ and length based formulations\end{tabular} &
  \begin{tabular}[c]{@{}l@{}}Captions treated as sequence\\ generation problem addressed \\  in target string\end{tabular} \\* \midrule
EMVC \cite{chang2022event} &
  \begin{tabular}[c]{@{}l@{}}Visual with Pre-trained TSN \\    Audio using VGGish pretrained \\ on AudioSet\end{tabular} &
  \begin{tabular}[c]{@{}l@{}}TEPs that incorporates audio cues\\ PtrNet for redundant events \\ and solve the problem of   \\ unbalanced recall and accuracy\end{tabular} &
  MHA + hRNN used for DCG \\* \midrule
\begin{tabular}[c]{@{}l@{}} \cite{jacob2022dense}\end{tabular} &
  Bi-SST + C + IIC (C3D features) &
  Bi-SST + C &
  LTA with Deep-LSTM (LSTM-D) \\* \midrule
\cite{zhang2022fine} &
  TSP features + I3D + Dlib &
  Followed baseline PDVC &
  Followed baseline PDVC \\* \midrule
\cite{dave2022hierarchical} &
  C3D for VFE &
  Bidirectional LSTM &
  LSTM decoder \\* \midrule
BIVT \cite{nishimura2022recipe} &
  Used pre-extracted features &
  \begin{tabular}[c]{@{}l@{}}TSN + Gumbel softmax resampling and \\ MIL-NCE with transformer   model\end{tabular} &
  Pretrained GloVe \\* \midrule
\cite{mittal2022savchoi} &
  \begin{tabular}[c]{@{}l@{}}VGGish + I3D + HOI for RGB features \\ + GloVe embeddings\end{tabular} &
  \begin{tabular}[c]{@{}l@{}}QAHOI (Query-based Anchors for \\ Human Object Interaction Detection)\end{tabular} &
  \begin{tabular}[c]{@{}l@{}}Dense Captions with NMS and \\ Text classification\end{tabular} \\* \midrule
SCC \cite{wang2022semantic} &
  R(2+1)D   and "SA-TSP + I3D + VGGish" &
  Predefined localized events by \cite{yu2021dense} &
  \begin{tabular}[c]{@{}l@{}}Language modeling +\\    Self-critical sequence training (SCST)\end{tabular} \\* \midrule
\cite{zhang2022unifying} &
  C3D and TSN for VFE &
  MEFM for localization&
  Pretraining with MLM and MVFR \\* \midrule
\cite{lu2022pic} &
  \begin{tabular}[c]{@{}l@{}}Pretrained concept detector followed \\ by multi-scale feature extractor\end{tabular} &
  Localization head with MLP backbone &
  Captioning head with LSTM backbone \\* \midrule
\begin{tabular}[c]{@{}l@{}} REASONER \\ \cite{liang2022visual} \end{tabular}&
  ResNet200 + BN-Inception &
  Causality-Aware Encoder &
  Cascaded-Reasoning Decoder \\* \midrule
PPVC \cite{choi2022parallel} &
  C3D for VFE &
  \begin{tabular}[c]{@{}l@{}}Transformer decoder without \\ self-attention block but \\ leverage cross-attention and CNN\end{tabular} &
  \begin{tabular}[c]{@{}l@{}}Transformer decoder with multi-stack \\ cross attention\end{tabular} \\* \midrule
VLCAP \cite{yamazaki2022vlcap} &
  C3D + CLIP for VFE &
  Pre-extracted set of events &
  \begin{tabular}[c]{@{}l@{}}Unified encoder-decoder with \\ transformer backbone\end{tabular} \\* \midrule
VLTint \cite{yamazaki2022vltint} &
  \begin{tabular}[c]{@{}l@{}}3D-CNN network for visual and \\ CLIP for linguistic features\end{tabular} &
  Pre-extracted set of events &
  \begin{tabular}[c]{@{}l@{}}Unified encoder-decoder called\\ Transformer-in-Transformer (TinT)\end{tabular} \\* \midrule
AMT \cite{yu2021accelerated} &
  Does not extract features &
  \begin{tabular}[c]{@{}l@{}}Lightweight proposal with no anchors \\ and a local attention method\end{tabular} &
  \begin{tabular}[c]{@{}l@{}}Single-shot feature blocking technique \\ and a standard attention mechanism\end{tabular} \\* \midrule
SC-Net \cite{aafaq2021cross} &
  Semantic contextualization &
  Event descriptor using SFT and EPN &
  Event subsequent captioning network \\* \midrule
VSEC \cite{dang2021dense} &
  \begin{tabular}[c]{@{}l@{}}Segment-sentence pair information to  \\  learn expressive features using VSEC\end{tabular} &
  \begin{tabular}[c]{@{}l@{}}Transformer encoder with proposal \\ bounding operation\end{tabular} &
  Decoder network masked transformer \\* \midrule
EA-DVC \cite{lu2021environment} &
  Used pre-extracted features &
  \begin{tabular}[c]{@{}l@{}}Event sequence compression with  \\ transformer to learn event dependency\end{tabular} &
  Transformer decoder \\* \midrule
\cite{estevam2021dense} &
  \begin{tabular}[c]{@{}l@{}} I3D + mini batch k-means + GloVe \\ used for VFE \end{tabular} &
  BMT for localization&
  \begin{tabular}[c]{@{}l@{}}Unsupervised semantic VT and \\ language generator trained on \\ GT events and sentences\end{tabular} \\* \midrule
HAED \cite{yu2021dense} &
  \begin{tabular}[c]{@{}l@{}}Visuals - C3D + TSN\\    Object features - Faster R-CNN\end{tabular} &
  SST + ESGN &
  Hierarchical decoder with text-attention \\* \midrule
\begin{tabular}[c]{@{}l@{}} DVCflow \\ \cite{yan2021dvcflow} \end{tabular}&
  \begin{tabular}[c]{@{}l@{}}Local and global features extraction using \\ MART \cite{lei2020mart} and \cite{xiong2018move} \\ to combat cross-modal information flow\end{tabular} &
  No Event Proposal module &
  BERT inspired language model \\* \midrule
DVC-Net \cite{lee2021dvc} &
  \begin{tabular}[c]{@{}l@{}}ResNet and C3D based \\ Visual Feature Extraction Network (VFEN)\end{tabular} &
  Bidirectional LSTM based TEPN &
  CGN with LSTM and context gating \\* \midrule
SME-DVC \cite{scherer2022semantic} &
  \begin{tabular}[c]{@{}l@{}}Pre-trained TSN as Masked Transformer \\ \cite{zhou2018end}\end{tabular} &
  \begin{tabular}[c]{@{}l@{}}Masked Transformer \cite{zhou2018end} \\ and PDVC \cite{wang2021end}\end{tabular} &
  \begin{tabular}[c]{@{}l@{}}Masked Transformer \cite{zhou2018end} \\ and PDVC \cite{wang2021end}\end{tabular} \\* \midrule
\begin{tabular}[c]{@{}l@{}}PDVC\\    \cite{wang2021end}\end{tabular} &
  \begin{tabular}[c]{@{}l@{}}Transformer encoder with CNN backbone \end{tabular} &
  SST + ESGN &
  \begin{tabular}[c]{@{}l@{}}Lightweight and standard captioning \\  head with LSTM backbone\end{tabular} \\* \midrule
\cite{kanani2021global} &
  ResNet and R3D &
  Adversarial Interface in Global proposals &
  LSTM decoder + Sentence discriminators \\* \midrule
\cite{boran2021leveraging} &
  ResNext-101 model + ImageNet features &
  Event clips treated as images &
  \begin{tabular}[c]{@{}l@{}}Auxiliary caption information and \\ LSTM decoder\end{tabular} \\* \midrule
\cite{chen2021towards} &
  C3D features &
  \begin{tabular}[c]{@{}l@{}}Sentence localizer with frame and word \\ modeling  MIL + ISAB   \\ \end{tabular} &
  Single-layer bidirectional LSTM \\* \midrule
\begin{tabular}[c]{@{}l@{}} KDPG-DVC \\ \cite{wu2021weakly} \end{tabular}&
  \begin{tabular}[c]{@{}l@{}}Modality features - Self-attention\\   Video - Residual learning\\    Text - BERT\end{tabular} &
  \begin{tabular}[c]{@{}l@{}}Knowledge distillation-based proposal \\ generation (KDPG) and COOT-based \\ \cite{ging2020coot} PCM\end{tabular} &
  Attention-LSTM network \\* \midrule
\begin{tabular}[c]{@{}l@{}}BMT\\    \cite{iashin2020better}\end{tabular} &
  \begin{tabular}[c]{@{}l@{}}Audio - VGGish\\    Visuals -I3D\\    Caption tokens - GloVe\end{tabular} &
  \begin{tabular}[c]{@{}l@{}}Bi-modal multi-headed proposal\\    generator\end{tabular} &
  \begin{tabular}[c]{@{}l@{}}Transformer based \\    Bi-modal decoder\end{tabular} \\* \midrule
\cite{suin2020efficient} &
  Standard feature extractor CNN &
  \begin{tabular}[c]{@{}l@{}}Frame Selection Network with\\    Guide Network Reinforcement learning\end{tabular} &
  \begin{tabular}[c]{@{}l@{}}Caption decoder inspired with \\ masked transformer and reinforcement\end{tabular} \\* \midrule
GPaS \cite{zhang2020dense} &
  Does not extract features &
  Bi-AFCG &
  GCN-enhanced summarization framework \\* \midrule
\cite{wang2020dense} &
  Features treated at frame-level using TSN &
  \begin{tabular}[c]{@{}l@{}}Dense Boundary Generator (DBG) \cite{lin2020fast} \\ and ESSN \\ \cite{chen2019activitynet}\end{tabular} &
  \begin{tabular}[c]{@{}l@{}}Temporal-semantic relation module  \\ (TSRM) using \cite{wang2020dense} and\\ Gated hierarchical RNN using CMG\end{tabular} \\* \midrule
\begin{tabular}[c]{@{}l@{}}TL-NMS \\ \cite{wang2020event}\end{tabular} &
  C3D feature extracted&
  SST for TEL &
  LSTM \\* \midrule
MDVC \cite{iashin2020multi} &
  C3D + VGGish + I3D &
  \begin{tabular}[c]{@{}l@{}}Bidirectional Single-stream temporal \\ action proposal (Bi-STT)\end{tabular} &
  Multimodal generator \\* \midrule
\cite{huang2020multimodal} &
  Compact 2D and S3D &
  Predefined video segments &
  Transformer decoder with attention \\* \midrule
\cite{song2020team} &
  Resnet200 + I3D + VGGish &
  Bi-directional temporal dependency &
  \begin{tabular}[c]{@{}l@{}}Predefined event captioning with  \\ GRU and LSTM backbone \cite{chen2019activitynet}\end{tabular} \\* \midrule
\cite{chen2019activitynet} &
  \begin{tabular}[c]{@{}l@{}}Base features - Resnet200 + I3D + VGGish \\    Object features - Faster R-CNN\\    Semantic features - Concept predictor\end{tabular} &
  \begin{tabular}[c]{@{}l@{}}Event Sequence Generation Network \\ (ESGN) with pointer network\end{tabular} &
  \begin{tabular}[c]{@{}l@{}}General Seq2seq caption  \\ generation using GT captions \\ and re-ranking\end{tabular} \\* \midrule
\begin{tabular}[c]{@{}l@{}}JEDDi-Net \\ \cite{xu2019joint}\end{tabular} &
 \begin{tabular}[c]{@{}l@{}} 3D CNN + \\ 3D Segment-of-Interest pooling \end{tabular} &
  R-C3D model &
  Hierarchical LSTM \\* \midrule
DaS \cite{zhang2019show} &
  C3D features &
  LSTM based TEL &
  2-stage LSTM for DCG \\* \midrule
SDVC \cite{mun2019streamlined} &
  C3D features &
  \begin{tabular}[c]{@{}l@{}}ESGN using Single-Stream Temporal \\ action proposals SST + PtrNet\end{tabular} &
  Sequential captioning with RNN \\* \midrule
SRN \cite{yang2019structured} &
  Available C3D+PCA features \cite{johnson2016densecap} &
  Latent distribution + stochastic attention  &
  LSTM + vector concatenation \\* \midrule
WLT \cite{rahman2019watch} &
  \begin{tabular}[c]{@{}l@{}}C3D for visual features and \\ audio feature processing using \\    MFCC + CQT + SoundNet features\end{tabular} &
  \begin{tabular}[c]{@{}l@{}}Non-negative matrix factorization (NMF) \\ for audio event detection and \\ cross-attention based sentence localizer\end{tabular} &
  Decoder with GRU backbone \\* \midrule
B-SST \cite{wang2018bidirectional} &
  C3D features &
  Forward/Backward Sequence Encoder &
  LSTM based decoder \\* \midrule
\cite{zhou2018end} &
  \begin{tabular}[c]{@{}l@{}}1-D features and optical \\ flow features using \cite{xiong2016cuhk}\end{tabular} &
  \begin{tabular}[c]{@{}l@{}}ProcNets inspired \\    anchor-offset mechanism\end{tabular} &
  Masked transformer \\* \midrule
\cite{li2018jointly} &
  3-D CNNs &
  Classification and regression &
  Attribute-augmented LSTM \\* \midrule
Mpp-Net \cite{wei2023mpp} &
  Hierarchical Temporal Spatial Summary &
  Multi-Perspective Perception Decoder &
  \begin{tabular}[c]{@{}l@{}}LSTM based multi-perspective \\ captioning generator\end{tabular} \\* \midrule
MFT \cite{xiong2018move} &
  Temporal Segmental Network (TSN) &
  CLIP with RNN &
  LSTM-based DCG \\* \midrule
WS-DEC \cite{duan2018weakly} &
  GRU for VFE &
  Attention based sentence localizer &
  Context aware caption generation \\* \midrule
\begin{tabular}[c]{@{}l@{}}RUC+CMU \\ \cite{chen2018ruc+}\end{tabular} &
  Resnet + I3D + VGGish + LSTM &
  Heuristic sliding window and ranking &
  \begin{tabular}[c]{@{}l@{}}VT, temporal attention and \\ topic guided caption\end{tabular} \\* \bottomrule
\end{longtable}
\end{singlespace}

Yunjie Zhang et al. used baseline PDVC for captioning video frames that properly localize essential points of interest. A hierarchical LSTM-based decoder optimizes captions using cross-entropy loss and reinforcement learning-based reward function \cite{zhang2022fine}. Li et al. used PDVC to generate dense captions from input video. Descriptions, timestamps, and confidence scores comprise the created caption set. The authors suggest \textbf{Non-Prompt Caption Suppression (NPCS)} to select high-quality and general prompt caption pairings from dense captions. The algorithm minimizes fast caption-ground truth interval overlap \cite{li2023generating}. Taichi Nishimura et al. used pre-trained GloVe for caption development. A hierarchical RNN-based decoder and video and audio features generate recipe-like captions \cite{nishimura2022recipe}. 

\textbf{Masked Transformers.}
Hiding specific tokens from the input sequence to block them from receiving attention in seq2seq problems is masking. The first masked transformer for DVC was proposed by Zhou et al. \cite{zhou2018end}. They generated captions by extracting visual and temporal characteristics from video frames using a two-stream network and a masked transformer-based decoder. Cross-entropy loss and reinforcement learning-based reward function help the model learn better word predictions. Following PDVC, SME-DVC has recently used the same approach \cite{scherer2022semantic}. Yu et al. used a single-shot feature masking strategy and an average attention mechanism \cite{yu2021accelerated}.

Suin et al. presented a masked transformer-inspired caption decoder and reinforcement system to generate output captions for the proposed event. The caption decoder uses a two-layer LSTM-based architecture. The first layer generates a series of tokens based on input features, and the second layer refines the sequence by attending to essential frames using a masked transformer-based attention mechanism \cite{suin2020efficient}. In \textbf{Visual-Semantic Embedding with Context (VSEC)} \cite{dang2021dense} Dang at al. applied baseline \cite{zhou2018end} masked transformer as they work for incomplete/missing video data. The proposed encoder-decoder architecture uses softmax and attention head similar to \cite{zhou2018end} masked transformer. 

\subsubsection{LSTM and RNN-based decoders}
Before transformers \cite{vaswani2017attention}, most NLP tasks, including caption generation, relied on RNNs and LSTMs. Recurrent neural networks like LSTM decoders generate output sequences from encoded input sequences. LSTM cells track context and generate outputs based on previous results and encoded input. 

\textbf{Bi-directional LSTM} One of the prior works for DVC, \textbf{Bi-SST} \cite{wang2018bidirectional} uses LSTM-based decoder to learn event dependencies and generates caption accordingly. A bidirectional LSTM considers video features from past and future contexts such as, \cite{madake2022dense}. Kanani et al. introduced a novel architecture called \textbf{Global Object Proposal (GOP)} that used an LSTM decoder and phrase discriminators to detect relevant objects in video frames and generate captions that focus on them \cite{kanani2021global}. Bidirectional LSTM used by \cite{wang2023learning} increases the expressive ability of the model using an \textbf{Event-to-Text Generation (ETG)} network. The model aims to guess the event occurrence time and how confident it is using MLP. Later, a lightweight LSTM decoder generates sentences. \textbf{Graph-based Partition-and-Summarization (GPaS)} \cite{zhang2020dense} employed GCN-enhanced summarization by leveraging graphed nodes and edges of video frames. LSTM combined with auxiliary caption information and hybrid discriminator \cite{boran2021leveraging} also improves dense video captioning. \textbf{ Multi-Perspective Perception Network (Mpp-Net)} uses multi-level attention mechanism to integrate multi-perspective characteristics at multiple LSTM layers in decoder block \cite{ wei2023mpp}. Weak supervision combined with LSTM is another powerful DCG technique used in literature \cite{wu2021weakly, chen2021towards}

\textbf{Hierarchical decoders with attention or context gating} Hierarchical LSTM (hLSTM) architectures \cite{xu2019joint, wang2020event} are well suitable for the scenarios where we need to tackle the redundancy in generated captions or events set. For example, \cite{wang2020event} proposed a two-level hierarchical LSTM model called \textbf{TL-NMS} that utilizes non-maximum suppression to improve proposal and event quality. Similarly, the Division and Summarization framework proposed by \cite{zhang2019show} emphasizes the proposal of a single event once it exploits the semantic knowledge. A few models exploit the power of both hierarchical and bi-directional LSTM decoders, such as \cite{dave2022hierarchical}. The model uses end-to-end learning and LSTM at both the event proposal and captioning stages.  A linguistic prediction network, \textbf{SSRN} \cite{yang2019structured} encodes video and text. The video encoder captures frame temporal dynamics with a 3D CNN and LSTM network. The text encoder predicts stochastic variables representing future frame uncertainty using an LSTM network and vector concatenation.

\textbf{Recurrent Neural Network models} At the feature extraction stage, various techniques have been applied to generate feature vectors. Such as MHA \cite{chang2022event}, TSRM \cite{wang2020dense}, and 3D CNN \cite{li2023time} while they all accompany RNN for caption generation. Weak supervision combined with RNN decoder \cite{mun2019streamlined} and GRU decoder \cite{rahman2019watch} is also used to generate captions. The single-stage model \cite{mun2019streamlined} jointly detects events and generates captions using cross-entropy loss and reinforcement learning to optimize caption quality and diversity. The attention method refines caption proposals and trains the GRU decoder to maximize caption maximum likelihood estimation. The training requires no ground-truth captions and weak supervision \cite{rahman2019watch}.

\textbf{Pretraining models for Dense Video Captioning} Pretraining Dense Video Captioning models on a big dataset to learn a broad representation of visual features and language interpretation. These models are trained to predict labels or word sequences from input data.

Huang et al. proposed a Transformer decoder with attention, a multimodal pretraining technique for dense video captioning. This Transformer decoder synthesized words from input frames using an attention method \cite{huang2020multimodal}. Another semantic-aware pretraining for dense video captioning \cite{wang2022semantic} utilizes SCST for caption creation. Language modeling pretrains a transformer-based model on text and visuals. A transformer decoder sequence-to-sequence architecture generates captions. SCST maximizes a caption quality reward function to optimize the model during training .

Vid2Seq, a dense video captioning pretraining method by Yang et al., combines language modeling for visual and speech embeddings in the decoder block. A visual language model is pre-trained on a huge corpus of text and videos to learn a rich visual and textual information representation. A sequence-to-sequence architecture with a transformer decoder employs visual and speech embeddings to identify key elements in input frames and generate informative captions \cite{yang2023vid2seq}.

\textbf{Joint learning models} \cite{ xu2019joint, li2018jointly, zhang2022unifying} or end-to-end models \cite{zhu2022end, zhou2018end, wang2021end} can capture complex associations between visual and linguistic information in input video frames to generate coherent captions using joint learning. 

DVC is also performed end-to-end instead of following the traditional 'localize-and-describe' pattern. The first end-to-end framework, \cite{zhou2018end}, used a masked transformer for caption generation followed by CNNs for visual feature extraction. The multi-modal pre-trained model \cite{zhu2022end} suggested an end-to-end dense video captioning method that treats captions as a sequence creation problem in the target string. The encoder processes visual input, and the decoder generates captions using a sequence-to-sequence architecture. The decoder's target string guides the attention mechanism, which estimates each frame's importance for developing the caption's current word. Similarly, "Move forward and tell" \cite{xiong2018move} works progressively to perform a selection of events and caption generation simultaneously.

Aafaq et al. presented an event subsequent captioning network for cross-domain modality fusion dense video captioning. First, a weakly supervised event localization network creates candidate events. An event's subsequent captioning network uses visual features and prior captions to caption each potential event in the second stage. An LSTM-based decoder successively generates captions using visual cues and previously created words \cite{aafaq2021cross}. 

Xu et al. \cite{xu2019joint} used a hierarchical LSTM network to recognize and describe events in continuous video streams for caption generation. The hLSTM network is trained in an end-to-end manner, utilizing supervised and unsupervised objectives to provide relevant and coherent captions for identified events. JLDE \cite{li2018jointly} uses an attribute-augmented joint-learning based LSTM-based model to generate captions for candidate events in input video frames. An attribute-based LSTM decoder generates captions using visual characteristics and attribute information extracted by a visual encoder. 

 Nishimura et al. remodeled the general flow of event selection criteria and introduced an oracle event selector as a candidate for caption generation \cite{nishimura2022recipe}. The model then employs a transformer-based joint learning approach for caption generation against selected events. Qi Zhang et al. proposed a unifying architecture \cite{zhang2022unifying} to promote better inter-task association between event and caption generation. Masked Language Modeling (MLM) and Masked Video Feature Regression (MVFR) aim to pretrain a transformer-based model to learn a rich representation of visual and textual input in a unified manner. 
  \section{Trends in Publishing Research in DVC}

Around 40 percent of the studies in DVC used transformer-based architectures, while 13 percent leveraged LSTM and RNN architectures. Architectures that take advantage of two or more ML architectures in the training or testing phase, the multi-modal architectures, are used in around 13 percent of studies. Joint learning models exploit single or multi-modal architectures while jointly localizing events and describing them simultaneously. Nine percent of studies are using such models. Weakly supervised learning models are used in 6 percent of total studies. Three percent of studies still need to implement further extraction of knowledge from DVC's output. Lastly, techniques such as common sense reasoning, zero-shot learning, and GANs took part in 16 percent of the remaining studies.

 \small
\begin{longtable}[c]{l|ccccccc}
\caption{Performance comparison among all DVC models on two popular datasets. \textbf{ActivityNet Captions} and \textbf{YouCook2}. Models marked with \textbf{*} present the model performance results on ActivityNet Caps. Models marked with \textbf{‡}  present results on the YouCook dataset. All results reported with predicted captions are listed if otherwise explicitly mentioned with \textbf{(GT)} for Ground Truth captions.}
\label{tab:my-table}\\
\hline
\multicolumn{1}{c}{\textbf{Year}} &
  \textbf{Model Name} &
  \textbf{Framework} &
  \textbf{ROUGE} &
  \textbf{METEOR} &
  \textbf{CiDer} &
  \textbf{BLEU@4} &
  \textbf{BLEU@3} \\ \hline
\endfirsthead
\multicolumn{7}{c}%
{{\bfseries Table \thetable\ continued from previous page}} \\
\hline
\multicolumn{1}{c}{\textbf{Year}} &
  \textbf{Model Name} &
  \textbf{Framework} &
  \textbf{ROUGE} &
  \textbf{METEOR} &
  \textbf{CiDer} &
  \textbf{BLEU@4} &
  \textbf{BLEU@3} \\ \hline
\endhead
\hline
\endfoot
\endlastfoot

& DVCTM \cite{palivela2023dense}*       & Multimodal architecture          & -     & 10.54     & 30.62      & 1.80      & 4.13      \\
                         & DVCL \cite{qian2023dense}*    & Attention aware transformer                   & -     & 8.61      & 29.97      & 1.88      & -         \\
                         & EAMA-DVC \cite{wei2023enhancing}* &  Contrastive learning              & -     & 8.33      & -          & 1.87      & 3.75      \\
                         & ZeroTa \cite{jo2023zero}*  & Zero-shot learning                     & -     & 2.7       & 7.5        & -         & -         \\
                         & ZeroTa \cite{jo2023zero} ‡  & Zero-shot learning                    & -     & 2.1       & 4.9        & -         & -         \\
                         & SBS \cite{choi2023step}*  & Sequential captioning                      & -     & 9.05      & 27.92      & 1.08      & -         \\
                         & LTNet \cite{xu2023latent} & Topic-aware & - & 7.73 & 28.90 & 2.10 & -  \\
                         & MS-FTN\cite{10227555} & Fusion based transformer & - & 8.22 & - & 2.06 & - \\
                         & MRCap \cite{chen2023mrcap}* & Contrastive learning & - & 4.85 & 23.87 & 0.98 & - \\
                         & MRCap \cite{chen2023mrcap}‡ & Contrastive learning & - & 8.17 & 31.26 & 2.14 & - \\
                         
\multicolumn{1}{c|}{2023} & FMMF \cite{huang2023fusion}*   & Multimodal transformer                      & -     & 10.24     & 32.82      & 1.91      & 4.03      \\
                         & Mpp-Net \cite{wei2023mpp}* & Multi-perspective generator                     & -     & 16.01     & 29.35      & -         & 12.75     \\
                         & Mpp-Net \cite{wei2023mpp}‡  & Multi-perspective generator                    &       & 4.81      & 24.11      & 0.99      &           \\
                         & VidL \cite{wang2023learning}*  & Joint learning model                 & -     & 16.35     & 33.33      & -         & 11.70     \\
                         & VidL \cite{wang2023learning}‡ & Joint learning model                  &       & 5.01      & 26.52      & 1.04      &           \\
                         & CMCR \cite{han2023lightweight}*  & Cross-model attention               & -     & 10.09     & -          & 2.06      & 4.27      \\
                         & TFTD* \cite{li2023time}  & Recurrent transformer                       &       & 16.56     & 26.16      & 11.36     &           \\
                         & TFTD‡ \cite{li2023time}    & Recurrent transformer                   &       & 16.56     & 33.92      & 7.78      &           \\
                         & Vid2Seq \cite{yang2023vid2seq}*  & Multimodal endec              & -     & 17.0      & 28.0       & -         & -         \\
                         & Vid2Seq \cite{yang2023vid2seq}‡  & Multimodal endec             & -     & 24.0      & 50.1       & -         & -         \\ \hline
                         
                         & AAP-MIT \cite{prudviraj2022aap}*   & Memory transformer            & 33.10 & 17.55     & 28.32      & 13.20     & -         \\
                         & AAP-MIT \cite{prudviraj2022aap}‡   & Memory transformer             & 37.32 & 18.23     & 49.18      & 9.82      & 16.76     \\
                         & \cite{madake2022dense}   (MSDV) & LSTM architecture          & -     & 34.7      & -          & 75.0      & 77.8      \\
                         & \cite{lu2022pic}‡  & Encoder/decoder                             & -     & 21.56(GT) & 135.45(GT) & 13.91(GT) & -         \\
 &
  SME-DVC \cite{scherer2022semantic} & Post DVC &- &6.34 &- &0.63 &3.10 \\
                         & VSJM-Net \cite{aafaq2022dense}*  & Encoder/decoder               & 25.37 & 12.89     & 26.52      & 3.89      & 5.58      \\
                         & VSJM-Net \cite{aafaq2022dense}‡   & Encoder/decoder              & 10.51 & 4.31      & 9.07       & 1.09      &           \\
                         & PPVC \cite{choi2022parallel}* & Deformable transformer                  & -     & 7.91      & 23.02      & 1.68      & 3.58      \\
                         & PPVC \cite{choi2022parallel}‡    & Deformable transformer               & -     & 4.94      & 19.70      & 0.89      & -         \\
                         & PAM \cite{guo2022post}*         & Post DVC                &       & 15.78     & 23.72      & 10.31     &           \\
                         & \cite{zhu2022end}‡  & Pretrained transformer                            & 23.25 & 11.99     & 0.86       & 9.87      & -         \\
 &
  \cite{jacob2022dense} & Multimodal learning &
  19.20 &
  7.89 &
  11.01 &
  1.91 &
  4.01 \\
\multicolumn{1}{c|}{2022} & EMVC \cite{chang2022event}*  & RNN                   &       & 9.64      & 21.00      & 1.88      & 3.84      \\
                         & EMVC \cite{chang2022event}‡   & RNN                  & 10.49 & -         & -          & 0.96      & -         \\
                         & \cite{zhang2022fine}   (YouMakeup)  & Pretrained PDVC     & -     & 16.36     & 19.40      & 9.45      & -         \\
                         & \cite{dave2022hierarchical} & Hierarchical RNN                    & -     & 9.25      &            &           & -         \\
                         & BIVT \cite{nishimura2022recipe}‡ & Joint learning               & -     & 7.51      & 39.06      & 1.92      & -         \\
                         & \cite{mittal2022savchoi}*  & Pretrained transformer                     & -     & 16.36     &            & 9.45      & 10.91     \\
                         & SCC \cite{wang2022semantic}  & Semantic  pretraining                   & -     & 11.50     & 52.16 (GT) & -         & 3.10 (GT) \\
                         & \cite{zhang2022unifying}  & Masked language modeling                      & -     & 11.01     & 54.75 (GT) & -         & 2.90 (GT) \\
                         & REASONER \cite{liang2022visual}* & Encoder/decoder               & -     & 16.43     & 30.08      & -         & 12.45     \\
                         & VLCAP \cite{yamazaki2022vlcap}* & Transformer endec                & 35.00 & 17.48     & 31.29      & 13.38     & -         \\
                         & VLTinT \cite{yamazaki2022vltint}  & Transformer endec             & 36.56 & 17.97     & 31.13      & -         & 14.50     \\ \hline
                         & AMT \cite{yu2021accelerated}* & Masked transformer                  & -     & 5.82      & 10.87      & 1.20      & 2.42      \\
                         & AMT\cite{yu2021accelerated}‡   & Masked transformer                 & -     & 2.43      & 4.88       & -         & -         \\
                         & SC-Net \cite{aafaq2021cross}* & Joint learning                    & 22.32 & 10.93     & 14.68      & 2.47      & 5.21      \\
                         & SC-Net \cite{aafaq2021cross}‡   & Joint learning                & 10.28 & 4.02      & 8.30       & 1.04      &           \\
 &
  EA-DVC \cite{lu2021environment} & Lightweight DVC &
  - &
  11.79 &
  - &
  2.78 &
  6.04 \\
                         & VSEC \cite{dang2021dense}* & Semantic-aware MT                      & -     & 9.57 (GT) & 44.19 (GT) & 2.02 (GT) & 3.70 (GT) \\
                         & \cite{estevam2021dense}* & Transformer                         & 13.62 & 8.65      & 12.82      & 2.55      & 4.57      \\
                         & HAED \cite{yu2021dense}  & Hierarchical endec                       & -     & 9.74      & 28.09      & 1.98      & -         \\
                         & DVCflow \cite{yan2021dvcflow}* & Pretrained transformer                & -     & 17.38     & 23.66      & 10.21     &           \\
\multicolumn{1}{c|}{2021} & DVCflow \cite{yan2021dvcflow}‡ & Pretrained transformer                 & -     & 16.5      & 34.7       & 7.87      & -         \\
                         & DVC-Net \cite{lee2021dvc}* & RNN backbone                      & -     & -         & 15.80      & 1.26      & -         \\
                         & PDVC \cite{wang2021end}* & Deformable transformer                        & -     & 15.80     & 20.45      & 10.24     & -         \\
                         & \cite{kanani2021global}* & Reinforcement learning                        & -     & 16.36     & 19.40      & 9.45      & -         \\
                         & \cite{boran2021leveraging}* & GAN                    & -     & 14.39     & 18.51      &           & 9.39      \\
                         & EC-SL \cite{chen2021towards}*  & Weakly supervised biLSTM                 & 13.02 & 7.49      & 21.21      & 2.78      & 1.33      \\
                         & KDPG-DVC \cite{wu2021weakly}*  & Cross-modal                 & -     & 7.06      & 14.25      & 2.71      & 1.33      \\ \hline
                         & BMT \cite{iashin2020better}* & Pretrained transformer                    & -     & 8.44      & -          & 1.88      & 3.84      \\
                         & \cite{suin2020efficient}* & Masked transformer                       & -     & 6.21      & 13.82      & 1.35      & 2.87      \\
                         & GPaS\cite{zhang2020dense}*  & GCN                     & 21.30 & 11.04     & 28.20      & 1.53      & -         \\
                         & GPaS\cite{zhang2020dense}‡  & GCN                    & 27.97 & 12.20     & 41.44      & 1.64      & -         \\
                         & TL-NMS \cite{wang2020event}*   & Pretrained CNN                 & -     & 7.91      & 14.71      & 1.29      & -         \\
\multicolumn{1}{c|}{2020} & TL-NMS \cite{wang2020event}‡ & Pretrained CNN                   & -     & 3.70      & -          & -         & -         \\
                         & MDVC\cite{iashin2020multi}* & Multimodal transformer                    & -     & 7.31      & -          & 1.07      & 2.60      \\
                         & \cite{song2020team}*  & Contextual reasoning                           & -     & 11.28     & 14.03      & 5.32      & 2.91      \\
                         & MPDVC\cite{huang2020multimodal}   ViTT  & Pretraining with transformer        & 33.10 & 12.43     & 0.90       & -         & -         \\
                         & MPDVC\cite{huang2020multimodal}‡ & Pretraining with transformer                 & 39.03 & 18.32     & 3.80       & 12.04     &           \\
                         & \cite{wang2020dense}* & Hierarchical RNN                          & -     & 11.49     & 49.34      & 2.85      & -         \\ \hline
                         & \cite{chen2019activitynet}* & Contextual reasoning                     & -     & 9.90      & 56.52(GT)  & 4.59(GT)  & -         \\
                         & JEDDi-Net \cite{xu2019joint}* & Joint learning                   & 19.63 & 8.81      & 19.88      & 4.06      & 1.63      \\
     & DaS \cite{zhang2019show}* & LSTM                      & -     & 10.71     & 31.41      & -         & -         \\
\multicolumn{1}{c|}{2019} & SDVC \cite{mun2019streamlined}  & Reinforcement learning                 & -     & 8.82      & 30.68      & 0.93      & 2.94      \\
                         & SRN \cite{yang2019structured}*  & End-to-end SRN                & 22.98 & 9.54      & 15.81      & 5.19      & 2.63      \\
                         & WLT \cite{rahman2019watch}*  & Multimodal architecture                    & 9.60  & 4.78      & 10.53      & 1.69      & 0.82      \\ \hline
                         & Bidirectional SST\cite{wang2018bidirectional}* & Bidirectional LSTM & 19.10 & 9.60      & 12.68      & 2.30      & 4.41      \\
                         & DVC-MT \cite{zhou2018end} & Masked transformer                              & -     & 9.56      & -          & 2.23      & 4.47      \\
                         & \cite{li2018jointly} & Joint learning                             & -     & 6.93      & 13.21      & 0.74      & 2.27      \\
\multicolumn{1}{c|}{2018} & MFT \cite{xiong2018move}   & LSTM                     & 25.88 & 14.75     & -          & 13.52     & 8.45      \\
                         & WS-DEC \cite{duan2018weakly} & Weakly supervised model                   & 12.55 & 6.30      & 18.77      & 1.27      & 2.62      \\
                         & RUC+CMU \cite{chen2018ruc+}  & Topic-aware transformer                   & -     & 12.44     & 31.10      & 4.00      & -         \\ \hline
\end{longtable}

\section{Datasets and Evaluation Metrics}
Datasets and evaluation metrics used in the field of DVC are explored in this portion of the article. We develop an illustration showcasing the wide use of a specific dataset, enabling useful insights. Furthermore, we provide a brief overview of the evaluation findings for the various methods incorporated in this survey, followed by a tabular representation. 

\subsection{Datasets used for DVC}
Dense video-captioning datasets are different from regular video-captioning datasets. The task of simple video captioning models is to generate a single paragraph or sentence that describes the whole video, while dense video captioning aims at capturing events in the videos and then telling each event in a natural language sentence. For the DVC models to capture these events, DVC datasets are defined that comprise event details along with captions. Here, we carefully include and discuss only those datasets used by studies that claim to perform dense captioning of events sentence by sentence. We also summarize the details of each dataset in Table \ref{tab:dataset table} and \ref{tab:Dataset table 2}.

\textbf{MSR-VTT (Video To Text)} \cite{xu2016msr} contains around 10,000 videos as it is one of the pioneer datasets used for describing the videos in natural language sentences. The dataset is collected using 257 unique queries, with each question corresponding to 118 videos. The videos belong to around 20 categories/domains, including movies, music, sports, news, howto, and 15 others. Each video is about 20 seconds long on average and has multiple human-generated descriptions. The dataset uses a 65:30:5 split for training, testing, and validation sets. MSR-VTT can be leveraged for various downstream tasks in natural language processing (NLP) and computer vision, including video captioning, summarization, retrieval, and text-to-video synthesis. 

\textbf{MSVD (Microsoft Research Video Description)} \cite{chen2011collecting} is a dataset of short video clips that are annotated with one or more natural language sentences. It contains 1970 video clips, with an average duration of approximately 10 seconds, annotated with 41,408 sentences. The dataset covers various categories, such as sports, cooking, and music, and was designed to evaluate the performance of automatic video description systems. MSDV can be used for several downstream tasks, such as VQA, video retrieval, video captioning, zero-shot VQA, and zero-shot learning. The MSVD dataset has been widely used in video understanding and natural language processing research and has led to significant advances in video captioning and retrieval systems. 

\begin{longtable}{ccccccc}
\caption{Detail of state-of-the-art datasets used for Dense Video Captioning(DVC)}
\label{tab:dataset table}\\
\hline
\textbf{Dataset Name} &
  \textbf{\begin{tabular}[c]{@{}c@{}}Videos \\ Time (hrs.)\end{tabular}} &
  \textbf{Total Clips} &
  \textbf{\begin{tabular}[c]{@{}c@{}}Avg. Clip \\ Length\\ (sec)\end{tabular}} &
  \textbf{\begin{tabular}[c]{@{}c@{}}Sent. \\ Per Clip\\ (avg)\end{tabular}} &
  \textbf{\begin{tabular}[c]{@{}c@{}}Total No. \\ of   sent.\end{tabular}} &
  \textbf{\begin{tabular}[c]{@{}c@{}}Avg. sent.   \\ Length\end{tabular}} \\ \hline
\endfirsthead
\multicolumn{7}{c}%
{{\bfseries Table \thetable\ continued from previous page}} \\
\hline
\textbf{Dataset Name} &
  \textbf{\begin{tabular}[c]{@{}c@{}}Videos \\ Time (hrs.)\end{tabular}} &
  \textbf{Total Clips} &
  \textbf{\begin{tabular}[c]{@{}c@{}}Avg. Clip \\ Length\\ (sec)\end{tabular}} &
  \textbf{\begin{tabular}[c]{@{}c@{}}Sent. \\ Per Clip\\ (avg)\end{tabular}} &
  \textbf{\begin{tabular}[c]{@{}c@{}}Total No. \\ of   sent.\end{tabular}} &
  \textbf{\begin{tabular}[c]{@{}c@{}}Avg. sent.   \\ Length\end{tabular}} \\ \hline
\endhead
\hline
\endfoot
\endlastfoot
MSR-VTT \cite{xu2016msr}                      & 41.2  & 10k    & 10-30 & 20   & 200k   & 290   \\
MSVD \cite{chen2011collecting}                & 4.13  & 1.97k  & 10    & 41   & 120k   & ~85k  \\
YouCook2 \cite{zhou2018towards}               & 175.6 & 2k     & 315   & 8.3  & 16,600 & 13.2  \\
VATEX -en \cite{wang2019vatex}                & 1300  & 41,250 & 20    & 10   & 826k   & 15.23 \\
ActivityNet Captions \cite{krishna2017dense} & 849   & 20k    & 180   & 3.65 & 100k   & 13.48 \\
YouMakeup \cite{wang2019youmakeup}            & 421   & 2800   & 9     & 10.9 & 30,636 & -     \\
ViTT \cite{huang2020multimodal}                                         & -     & 8000   & -     & -    & 56,027 & 2.97  \\ \hline
\end{longtable}

\textbf{YouCook2}\cite{zhou2018towards} is a vast YouTube cooking videos dataset, one of the most popular datasets for the task of DVC after ActivityNet Captions. It contains around 2,000 cooking videos with more than 89,000 video segments. The dataset includes a variety of recipes from multiple cuisines, and the videos are captured from a first-person perspective, allowing for a natural and immersive experience. Each video is accompanied by a recipe text, which includes a list of ingredients and instructions for preparing the dish. The dataset can be used for various downstream tasks, including video summarization, video captioning, and recipe generation. Video summarization aims to shorten the original video by selecting and stitching together the most important segments. Recipe generation is a novel task that generates recipes from video and text data.

\textbf{VATEX} \cite{ wang2019vatex} is a multilingual video-and-text open domain dataset and is linguistically richer than MSR-VTT. It consists of 41,250 video clips with a total duration of 155 hours, collected from YouTube. Videos contain 600 unique actions. Each video is associated with multiple language descriptions, including English, Chinese, and Spanish, that are not translations of each other. There are 825,124 human translations, with an average of 20 translations per video. The dataset presents two tasks: Multilingual machine translation and video captioning. The paper also discusses the dataset's applicability to perform video retrieval and zero-shot few-shot learning capacity. 

\textbf{ActivityNet Captions} \cite{ krishna2017dense} is a large-scale and first DVC dataset containing over 20,000 videos and around 100,000 captions with start and end time stamps. The videos in the dataset come from diverse sources, including YouTube, Flickr, and Vimeo, and contain a wide range of activities, like cooking, dancing, sports, and more. The captions are written by human annotators and are designed to capture the salient aspects of the video.
The dataset can be used for various downstream tasks, including video captioning, summarization, temporal event localization, and activity recognition. The dataset has been used in several research studies to develop state-of-the-art methods for video captioning.

\begin{longtable}[c]{cccccc}
\caption{Original train, test and validation split of DVC datasets. Source defines the origin from where the dataset is collected along with domain of that dataset. Datasets that contain localization as downstream task are marked with \checkmark.}
\label{tab:Dataset table 2}\\
\hline
\textbf{Dataset Name}                          & \textbf{Source} & \textbf{Training Set} & \textbf{Validation Set} & \textbf{Testing Set} & \textbf{Localization} \\ \hline
\endfirsthead
\multicolumn{6}{c}%
{{\bfseries Table \thetable\ continued from previous page}} \\
\hline
\textbf{Dataset Name}                          & \textbf{Source} & \textbf{Training Set} & \textbf{Validation Set} & \textbf{Testing Set} & \textbf{Localization} \\ \hline
\endhead
\hline
\endfoot
\endlastfoot
MSR-VTT \cite{xu2016msr}           & Open (YT)    & 6,513  & 497  & 2,990  &  - \\
MSVD \cite{chen2011collecting} & Open (YT)    & 1200   & 100  & 670    &  -  \\
YouCook2 \cite{   zhou2018towards} & Cooking (YT) & 1,333  & 457  & 210    & \checkmark \\
VATEX -en \cite{wang2019vatex}     & Open         & 25,991 & 3000 & 12,278 &  - \\
ActivityNet Captions  \cite{krishna2017dense} & Human act. (YT) & 10009                 & 4925                    & 5044                 & \checkmark                     \\
YouMakeup \cite{wang2019youmakeup} & Makeup (YT)  & 1680   & 280  & 840    & \checkmark \\
ViTT \cite{ huang2020multimodal}   & Open         & -      & -    & -      & \checkmark \\ \hline
\end{longtable}

\textbf{YouMakeup} dataset \cite{wang2019youmakeup} contains 2,800 makeup videos for semantic comprehension and casualty action reasoning. Since different makeup steps share the same knowledge of the background, the fine-grained information challenge can be leveraged. The spatial face area and timestamp information make the dataset suitable for incorporating fine-grained information in the DVC task. The dataset has not been extensively implemented yet. 

\textbf{ViTT - Video Timeline Tags} \cite{huang2020multimodal} dataset is mainly curated to target the topic distribution in the wild. It comprises 8k videos, mainly of cooking and a few other domains collected from YouTube 8M dataset. ViTT is a pretraining dataset designed for DVC tasks and targets DVC using ASR+Vid→CAP. It consists of around 88k segments, as YouCook2 comprises only 11.5k. The dataset has not been widely used yet.   
\subsection{Evaluation Metrics}
Results for DVC in this survey are presented using four widely used evaluation metrics: \\
\textbf{(1) BLEU (Bilingual Evaluation Understudy):} BLEU measures the overlap of n-gram values between the generated and reference texts, which ranges from worst(0) to best(1). \\
\textbf{(2) METEOR (Metric for Evaluation of Translation with Explicit ORdering): } It considers stemming, synonyms, and order of words also ranging from worst(0) to best(1). \\ 
\textbf{(3) CIDEr (Consensus-based Image Description Evaluation):} Starting from the worst at 0, CIDEr targets the best higher value by considering the n-gram occurrences following comparison between generated and given captions. \\
\textbf{(4) ROUGE-L (Recall-Oriented Understudy for Gisting Evaluation - Longest common subsequence):} It measures the overall longest common subsequence between generated and given text by interpreting values as overlaps.

\begin{longtable}{cm{1.5in}ccc}
\caption{Collection of popular DVC datasets along with benchmark studies (2018-2023)}
\label{tab:benchmark table for datasets}\\
\hline
\textbf{Dataset} &
  \textbf{Representation} &
  \textbf{Annotated by} &
  \textbf{Videos source} &
  \multicolumn{1}{c}{\textbf{Benchmarks}} \\ \hline
\endfirsthead
\multicolumn{5}{c}%
{{\bfseries Table \thetable\ continued from previous page}} \\
\hline
\textbf{Dataset} &
  \textbf{Representation} &
  \textbf{Annotated by} &
  \textbf{Videos source} &
  \multicolumn{1}{c}{\textbf{Benchmarks}} \\ \hline
\endhead
\begin{tabular}[c]{@{}c@{}}ActivityNet \\ Captions\end{tabular} &
 \includegraphics[width=1.5in]{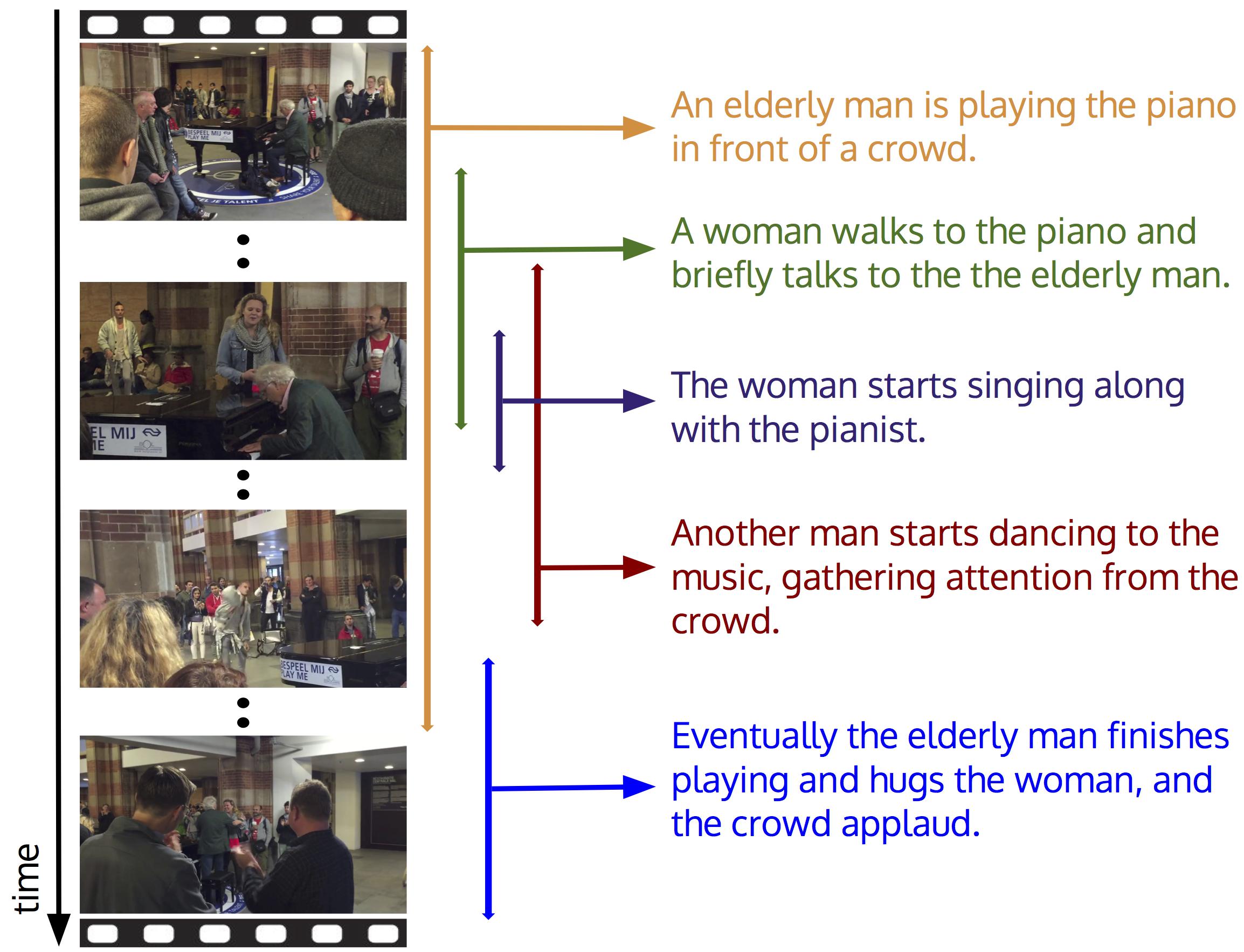}  &
  \begin{tabular}[c]{@{}c@{}}Amazon Mechanical \\ Turk (AMT)\end{tabular} &
  YouTube &  \begin{tabular}[c]{@{}l@{}}\cite{ qian2023dense,   estevam2021dense, palivela2023dense, zhang2020dense, aafaq2022dense,   yu2021dense} \\ \cite{jacob2022dense, wang2020dense, yan2021dvcflow, lee2021dvc,   zhu2022end, zhou2018end} \\ \cite{wang2021end, wang2020event, lu2021environment, chang2022event, wei2023enhancing, huang2023fusion} \\ \cite{li2023generating,   kanani2021global, dave2022hierarchical, li2018jointly, wang2023learning} \\  \cite{boran2021leveraging, han2023lightweight,   wei2023mpp, xiong2018move, iashin2020multi} \\ \cite{choi2022parallel, guo2022post, jung2023retrieval, mittal2022savchoi, zhang2019show, scherer2022semantic} \\ \cite{wang2022semantic, mun2019streamlined, choi2023step, yang2019structured,   song2020team, li2023time} \\ \cite{chen2021towards,  zhang2022unifying, yang2023vid2seq, yamazaki2022vlcap, yamazaki2022vltint, rahman2019watch} \\ \cite{duan2018weakly,   wu2021weakly, chen2018ruc+, 10227555, chen2023mrcap} \end{tabular} 
   \\ \hline
YouCook2  & \includegraphics[width=1.5in]{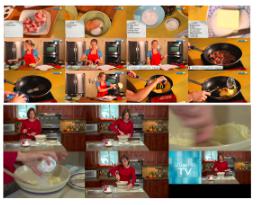}  & Human annotators                                                                   & YouTube              &  \begin{tabular}[c]{@{}l@{}}\cite{prudviraj2022aap,   yu2021accelerated, aafaq2021cross, qian2023dense, zhang2020dense,   yan2021dvcflow} \\ \cite{zhu2022end, zhou2018end, wang2021end, wang2020event,   chang2022event, huang2023fusion} \\ \cite{wang2023learning, wei2023mpp, huang2020multimodal, nishimura2022recipe, chen2023mrcap} \\ \cite{choi2023step, li2023time,   yang2023vid2seq, yamazaki2022vlcap, yamazaki2022vltint, jo2023zero}\end{tabular} \\ \hline
YouMakeup & \includegraphics[width=1.5in]{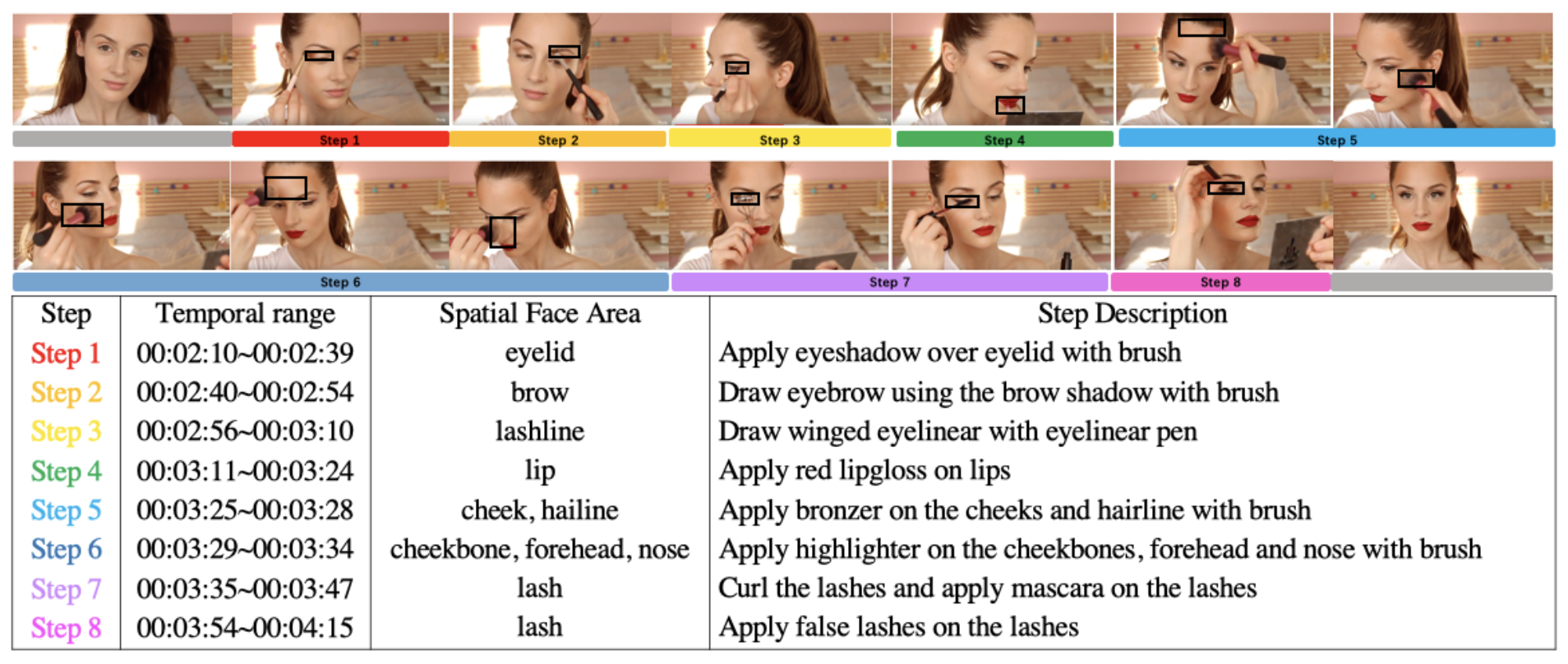}  & \begin{tabular}[c]{@{}c@{}}Video Subtitles \\ aligned with timestamps\end{tabular} & YouTube              &  \cite{lu2022pic,   zhang2022fine, wang2023learning, wang2019youmakeup}   \\ \hline
MSVD      & \includegraphics[width=1.5in]{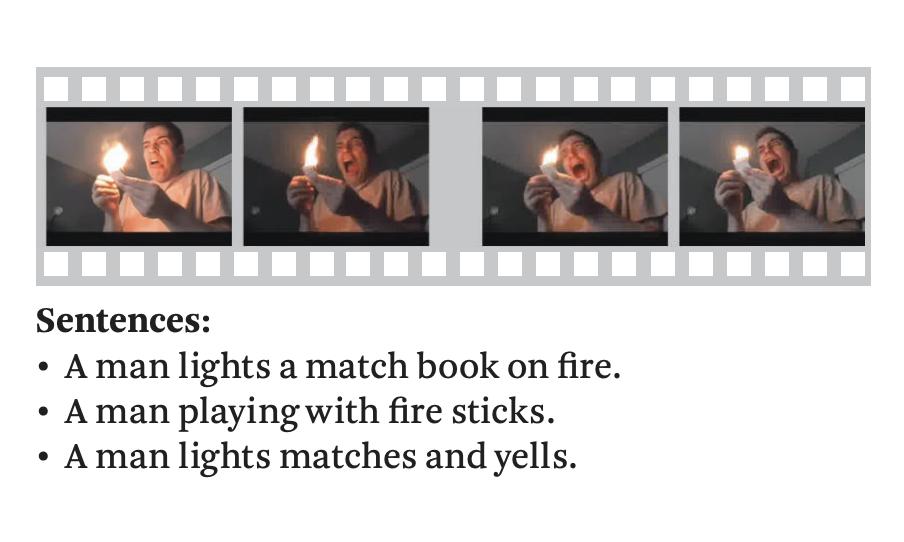}  & \begin{tabular}[c]{@{}c@{}}Amazon Mechanical \\ Turk (AMT)\end{tabular}            & YouTube              & \cite{madake2022dense,   yang2023vid2seq}  \\ \hline
MSR-VTT & \includegraphics[width=1.5in]{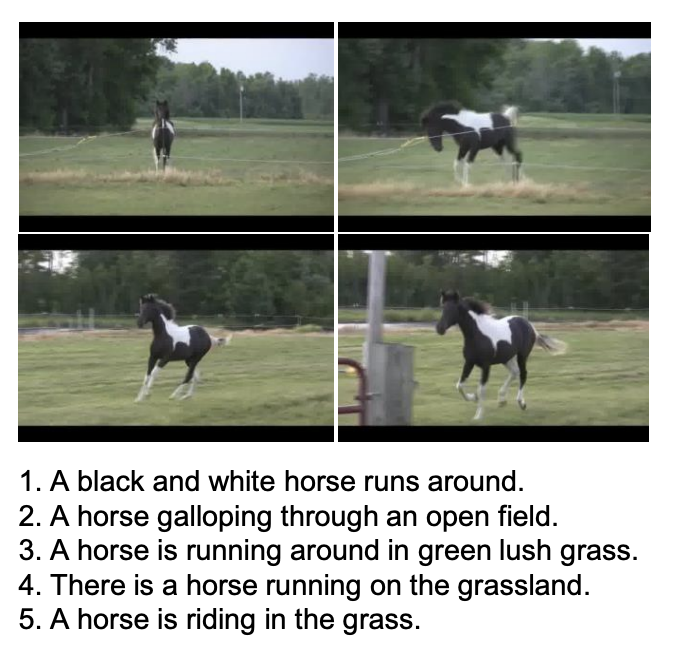} 
   &
  \begin{tabular}[c]{@{}c@{}}Amazon Mechanical\\  Turk (AMT)\end{tabular} &
  \begin{tabular}[c]{@{}c@{}}Commercial video \\ search engines\end{tabular} & \cite{lee2021dvc, huang2023fusion, zhang2022unifying, yang2023vid2seq}
   \\ \hline
ViTT      &  \includegraphics[width=1.5in]{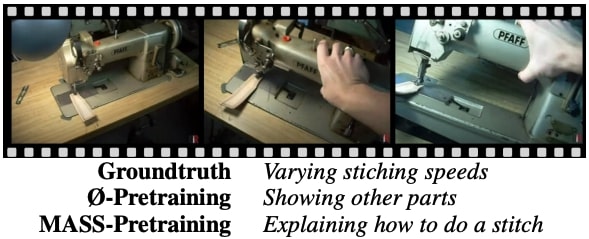}  & Human annotators                                                                   & YouTube 8M-Dataset   & \cite{   zhu2022end, huang2020multimodal, yang2023vid2seq}       \\ \hline
Vatex     & \includegraphics[width=1.5in]{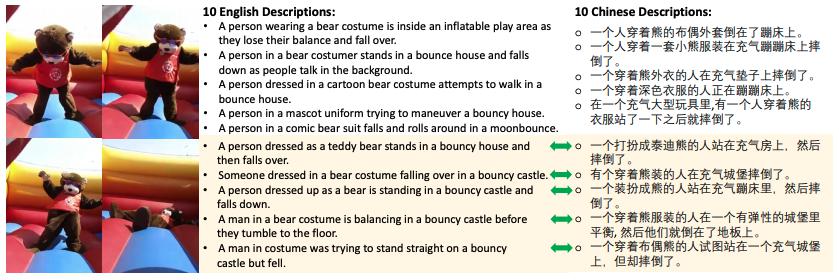}  & \begin{tabular}[c]{@{}c@{}}Amazon Mechanical\\  Turk (AMT)\end{tabular}            & Kinetics-600 dataset & 
 \cite{zhang2022unifying}  \\ \hline
\end{longtable}

\begin{figure}[H]
    \centering
    \includegraphics[width=350px]{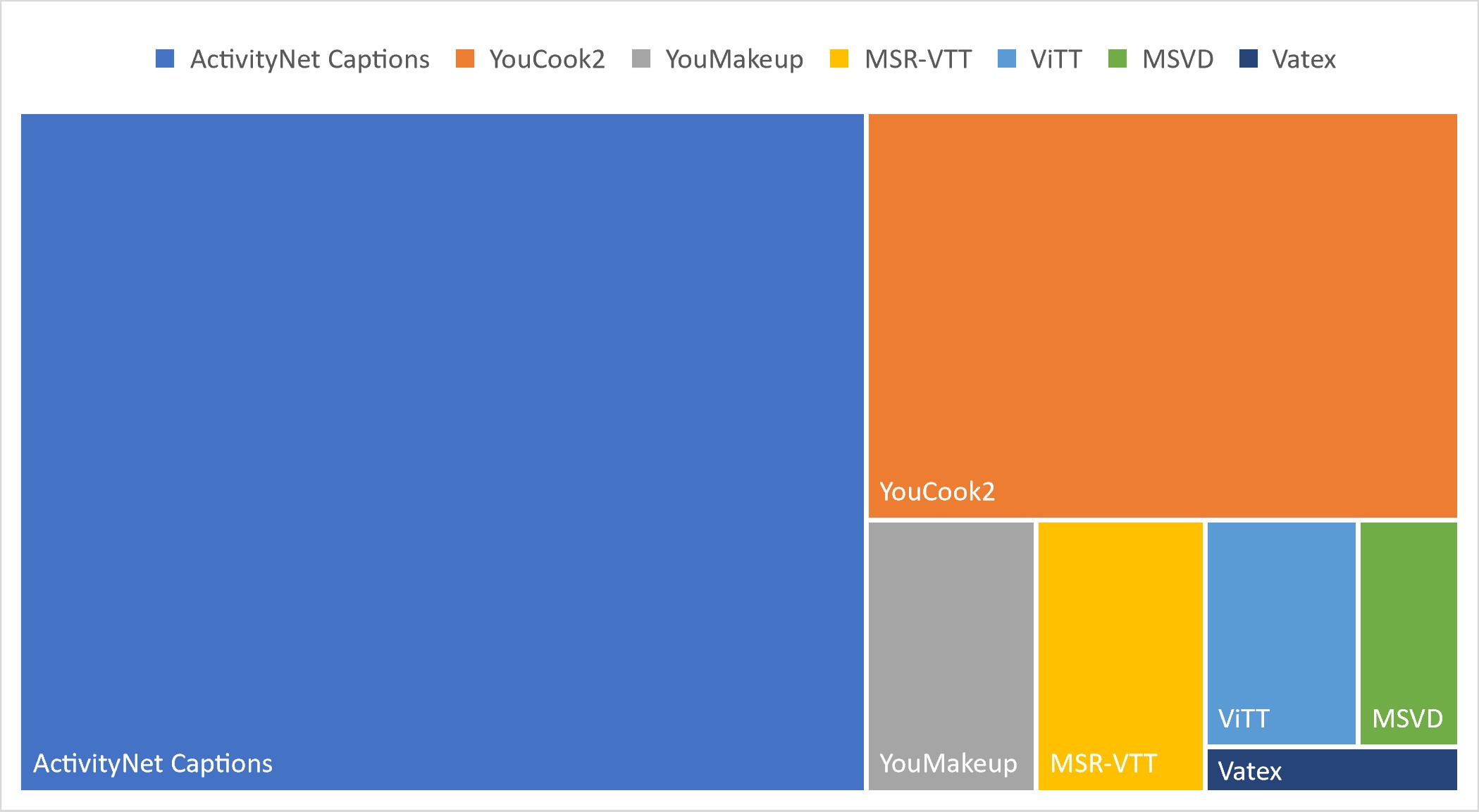}
    \caption{Treemap presenting prevalent use of various DVC datasets}
    \label{fig:enter-label}
\end{figure}

\section{Challenges and Future Directions}
Successes in image captioning tasks have inevitably led to an increasing number of interests in video captioning tasks. Since videos are rich in diversified content that leads to a series of events in a video, it is difficult to describe a video in a single sentence. Segmentation of the video is an intuitive method to identify valuable events in a video, but it only sometimes leads to better event detection. For example, a video using a transition effect during a single event does not mean the start of a new event in the video. Moreover, events in a video are mostly interrelated and demand reasonable coherence in the description.

\begin{figure}[H]
    \centering
    \includegraphics[width=400px]{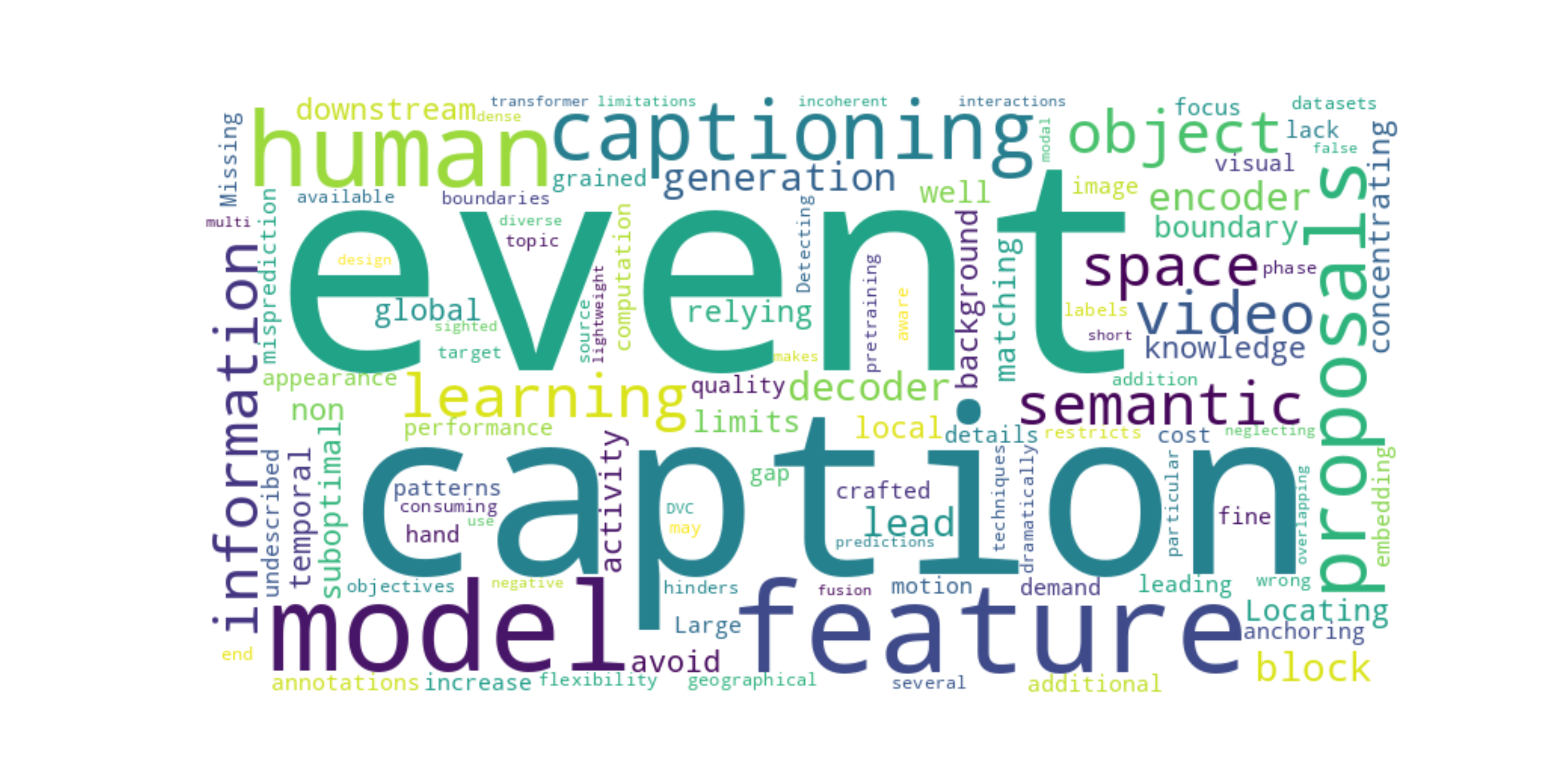}
    \caption{Word cloud reflecting challenges in dense video captioning}
    \label{fig:DVC_issues_word_cloud}
\end{figure}

\pagebreak
Here are some of the principal challenges from lower to higher severity that DVC encounters: 
\begin{itemize}
\item Locating temporal boundary of an event while concentrating on the local (human, non-human objects) as well as global (background knowledge) features to avoid suboptimal matching \cite{prudviraj2022aap,yamazaki2022vltint,yamazaki2022vlcap}
\item Since captioning is a downstream activity, relying on the event proposals limits its learning performance \cite{9710686}
\item The hand-crafted anchoring patterns for event proposals increase the cost of computation \cite{9710686}
\item Missing details, lack of focus on fine-grained information, undescribed appearance, and motion information, all leading to event/object misprediction in videos \cite{8744407,prudviraj2022aap,zhang2022fine,wei2023mpp} 
\item Large semantic gap between visual feature space and semantic embedding space \cite{ 10.1145/3550276,yan2021dvcflow}
\item The demand for additional annotations, like quality image captions and geographical labels, restricts the flexibility of available techniques, and the addition of several dataset objectives hinders the pretraining phase and is dramatically source-consuming \cite{wang2022git,suin2020efficient}
\item If the model is not aware of the 'topic of the target' in a particular video, it may end up in the wrong caption generation \cite{Deng_2021_CVPR,palivela2023dense}
\item Detecting boundaries of events while neglecting overlapping events leads to false event predictions and hence negative captions \cite{aafaq2022dense}
\item No-lightweight-design of dense captioning models \cite{lu2021environment}
\item Learning multimodal interactions and diverse feature fusion \cite{wei2023enhancing,chang2022event}
\item Most of the models use transformer encoder-decoder block for DVC; any limitations in the encoder block make the decoder short-sighted and lead to incoherent caption generation \cite{wei2023mpp,wei2023enhancing}
\end{itemize}

\subsection{Future Directions} \label{future directions}
Here, we discuss the potential future directions, concentrating on how far the field of dense video captioning will go in the years to come. 

\textbf{1. Leveraging Large-scale Pre-trained Models for Zero-shot Learning:} Developing the algorithms' capacity to adapt to complex and particular fields is an essential objective. It is crucial to look into methods for fine-tuning these pre-trained models so they can accurately capture domain-specific complexities. One such stepping stone could be zero-shot learning, as it aims to close the gap between seen and unseen classes. Furthermore, shifting the emphasis from single-modality comprehension to multimodal integration represents a fascinating direction. It may be possible to unlock improved cross-modal comprehension and encourage richer representations for zero-shot challenges by investigating approaches seamlessly combining visual, textual, and maybe aural channels within the framework of huge pre-trained models. Further research should be done on techniques for reducing biases and enhancing generalization in zero-shot settings as DVC advances.

\textbf{2. Topic-modeling for Improved Category-level Event-detection:} Meaningful captioning relies mainly on the issue of understanding and expressing events in videos. Researchers can find hidden concepts and patterns in videos using topic modeling strategies such as LDA and NFM. The system can offer more contextually detailed and logical descriptions by linking captions with certain topics, increasing the precision and detail in category-level event recognition. 

\textbf{3. Working with More Sophisticated Event-Proposal Models:} Dense captioning of events relies mostly on detecting events. The necessity to improve these models has recently been highlighted in many studies. More sophisticated event-proposal models that are computationally cost-effective can be proposed by utilizing deep neural architectures and advanced AI techniques. Moreover, refining event-proposal models through attention mechanisms, diversity loss, and reinforcement learning techniques will allow them to constantly focus on regions of interest and adapt to various video information. 

\textbf{4. Increasing Model Accuracy:} Another prominent future direction lies in the accuracy improvement of DVC models. Prominent benchmarks for accuracy measures are ROUGE-L, METEOR, SODA, BLEU, and CIDEr. Through persistent attempts to maximize these parameters and make sure that captions seem not solely fluent but also semantically linked with the visual material, it is clear that the emphasis is on improving model correctness. 

\textbf{5. Use of Knowledge graphs for increased event-caption relevance:} In the intricate realm of dense video captioning, incorporating knowledge graphs offers an appealing path for improving the relevance of event captions. Knowledge graphs provide extensive contextual data and relevant knowledge that may be used to improve the precision of event explanations. Knowledge graphs ensure the addition of domain-specific information, conceptual structures, and object properties. For instance, integrating data from the knowledge graph regarding an individual's occupation, location, or historical importance might enhance the recognition of a person engaging in a particular activity. This technique also has the potential to bridge the gap between visual content and external information to enhance the narrative concepts of videos. 

\textbf{6. Generating automatically annotated datasets:} Manually annotating large datasets is computationally expensive and laborious. On the other hand, the automatic generation of captions can be done using large-scale pre-trained models, computer vision techniques, and other NLP methods. Hence, future DVC datasets can leverage automatic annotations of captions to generate more datasets. For example, they produce captions automatically determined by the objects, situations, and activities observed in a video using object recognition systems.

\textbf{7. Revolutionizing Medical Insights:} As of our current knowledge, no dense video captioning model is available for medical video datasets. This can be another promising avenue in the realm of DVC. Medical video involves complex surgical instances, microscopic data, and diagnostic videos that can be densely captioned. Although annotating such datasets and training them for DVC is difficult, the potential benefits of such models are significant. They range from personalized individual training to automated surgical documentation.

\section{Conclusive Remarks}
This survey presented a comprehensive study of the Dense Video Captioning techniques. All the methods that came after the premise of DVC in the ActivityNet challenge are assembled in any survey for the first time. The survey is structured in the same way as a DVC technique's pipeline, i.e., (1) Video Feature Extraction, (2) Temporal Event Localization, and (Dense Caption Generation). We noticed that, not all the studies explicitly extracted video features, a few used pre-extracted features. Also, there is a large reuse of a few  outperforming temporal localization methods and transformers for decoding. After a thorough review of studies and the datasets, we summarized results for four evaluation metrics (BLEU, ROUGE-L, CIDEr, and METEOR). Datasets and evaluation metrics are also discussed in detail. Lastly, we identified significant challenges that DVC faces and how far the future of DVC can be seen. This review diligently examined the publishing trajectory from 2018 to October 2023, finding a significant and consistent rising trend in publication volumes. The steady increase highlights the field's ongoing growth and expansion of knowledge while also denoting more curiosity in it. 

Dense event detection and video captioning should be more broadly understood and combined correctly. Video Captioning does not deal with the events in the video separately; conversely, the core idea of \textit{dense} video captioning lies in treating each event individually. Most techniques start by using feature extraction to extract events from the videos. The most implied methods for feature extraction are C3D, VGGish, and I3D. Events are temporally localized using the extracted features or training and reinforcement learning on the dataset. Localization techniques can be broadly divided into proposal-based and proposal-free processes, and they rely on boundary-matching mechanisms and sliding windows. Many contemporary approaches continue to draw inspiration from these traditional methods despite being developed more than a decade ago. Transformer decoder block is the most practiced approach used for caption generation after localization of events. Whereas people still use variants of LSTMs and RNNs for DCG. For a comprehensive overview of VFE, TEL, and DCG, readers can refer to Table \ref{tab: Comprehensive overview of studies}.     
 
Around 80 percent of the studies have used the ActivityNet Captions challenge to evaluate their DVC model. Future DVC datasets are supposed to leverage timestamp information for events to qualify for suitable DVC datasets, as many researchers have highlighted its practicability. Apart from metrics discussed to evaluate video captioning for dense events, SODA is also widely used. The field of DVC is continuously evolving, with around 25 research methodologies published in the last year. Most of the research concerning dense caption generation is published in tier-one journals and conferences, which portrays the strong roots of the topic. In the years to come, DVC has many challenges to deal with, for example, amalgamation with the medical field, increasing the model's accuracy, coming up with more evaluation criteria, better pre-trained models, and others (section \ref{future directions}).

\begin{acks}
This work was supported by the Research Council of Norway Project (nanoAI ) 325741, H2020 Project(OrganVision) 964800, and VirtualStain(UiT) Cristin Project ID: 2061348.
\end{acks}

\bibliographystyle{ACM-Reference-Format}
\bibliography{sample-base}

\appendix

\end{document}